\pdfoutput=1
\pdfoutput=1
\pdfoutput=1
\pdfoutput=1
\pdfoutput=1
\documentclass[sigconf]{acmart}

\usepackage{booktabs} 
\usepackage{multirow}
\usepackage{fancyhdr}
\usepackage{amsmath}
\usepackage{amsthm}
\usepackage{url}
\usepackage{caption}
\usepackage[ruled,vlined]{algorithm2e}
\usepackage{graphicx,subfigure}
\usepackage{graphics}

\let\oldmaketitle\maketitle
\renewcommand{\maketitle}{%
  \oldmaketitle%
  \thispagestyle{fancy}}

\usepackage{hyperref}
\usepackage{cleveref}

\usepackage[hyphenbreaks]{breakurl}

\copyrightyear{2018}
\acmYear{2018}
\setcopyright{acmcopyright}
\acmConference[SIGIR '18]{The 41st International ACM SIGIR Conference on Research and Development in Information Retrieval}{July 8--12, 2018}{Ann Arbor, MI, USA}
\acmBooktitle{SIGIR '18: The 41st International ACM SIGIR Conference on Research and Development in Information Retrieval, July 8--12, 2018, Ann Arbor, MI, USA}
\acmPrice{15.00}
\acmDOI{10.1145/3209978.3210025}
\acmISBN{978-1-4503-5657-2/18/07}

\begin{document}

\title{Enhancing Person-Job Fit for Talent Recruitment:\\ An Ability-aware Neural Network Approach} \titlenote{This is an extended version of our SIGIR'18 paper.}

\author{Chuan Qin$^{1,2}$, Hengshu Zhu$^{2}$, Tong Xu$^{1,2}$, Chen Zhu$^2$, Liang Jiang$^1$, Enhong Chen$^1$, Hui Xiong$^{1,2,3,4}$}
\titlenote{Hui Xiong and Hengshu Zhu are corresponding authors.}
\affiliation{%
 \institution{$^1$Anhui Province Key Lab of Big Data Analysis and Application, University of Science and Technology of China\\
             $^2$Baidu Talent Intelligence Center, Baidu Inc. \ $^3$Business Intelligence Lab, Baidu Research\\
             $^4$National Engineering Laboratory of Deep Learning Technology an Application, China}
}
\affiliation{\{chuanqin0426, xionghui\}@gmail.com, \{zhuhengshu, zhuchen02\}@baidu.com}
\affiliation{jal@mail.ustc.edu.cn, \{tongxu, cheneh\}@ustc.edu.cn}

\begin{abstract}

The wide spread use of online recruitment services has led to information explosion in the job market. As a result, the recruiters  have to seek the intelligent ways for Person-Job Fit, which is the bridge for adapting the right job seekers to the right positions. Existing studies on Person-Job Fit have a focus on  measuring the matching degree between the talent qualification and the job requirements mainly based on the manual inspection of human resource experts despite of the subjective, incomplete, and inefficient nature of the human judgement. To this end, in this paper, we propose a novel end-to-end \textbf{A}bility-aware \textbf{P}erson-\textbf{J}ob \textbf{F}it \textbf{N}eural \textbf{N}etwork (APJFNN) model, which has a goal of reducing the dependence on manual labour and can provide better interpretation about the fitting results. The key idea is to exploit the rich information available at abundant historical job application data. Specifically, we propose a word-level semantic representation for both job requirements and job seekers' experiences based on Recurrent Neural Network (RNN). Along this line, four hierarchical ability-aware attention strategies are designed to measure the different importance of job requirements for semantic representation, as well as measuring the different contribution of each job experience to a specific ability requirement. Finally, extensive experiments on a large-scale real-world data set clearly validate the effectiveness and interpretability of the APJFNN framework compared with several baselines.


\end{abstract}



\maketitle

\section{Introduction}
The rapid development of online recruitment platforms, such as LinkedIn and Lagou, has enabled the new paradigm for talent recruitment. For instance, in 2017, there are \emph{467 million} users and \emph{3 million} active job listings in LinkedIn from about \emph{200} countries and territories all over the world~\cite{linkedin500m}. While popular online recruitment services provide more convenient channels for both employers and job seekers, it also comes the challenge of \textbf{Person-Job Fit} due to information explosion. According to the report~\cite{shrm2016}, the recruiters now need about 42 days and \$4,000 dollars in average for locking a suitable employee~\cite{shrm2016}. Clearly, more effective techniques are urgently required  for the Person-Job Fit task, which targets at measuring the matching degree between the talent qualification and the job requirements.

\begin{figure}
  \includegraphics[width=\linewidth]{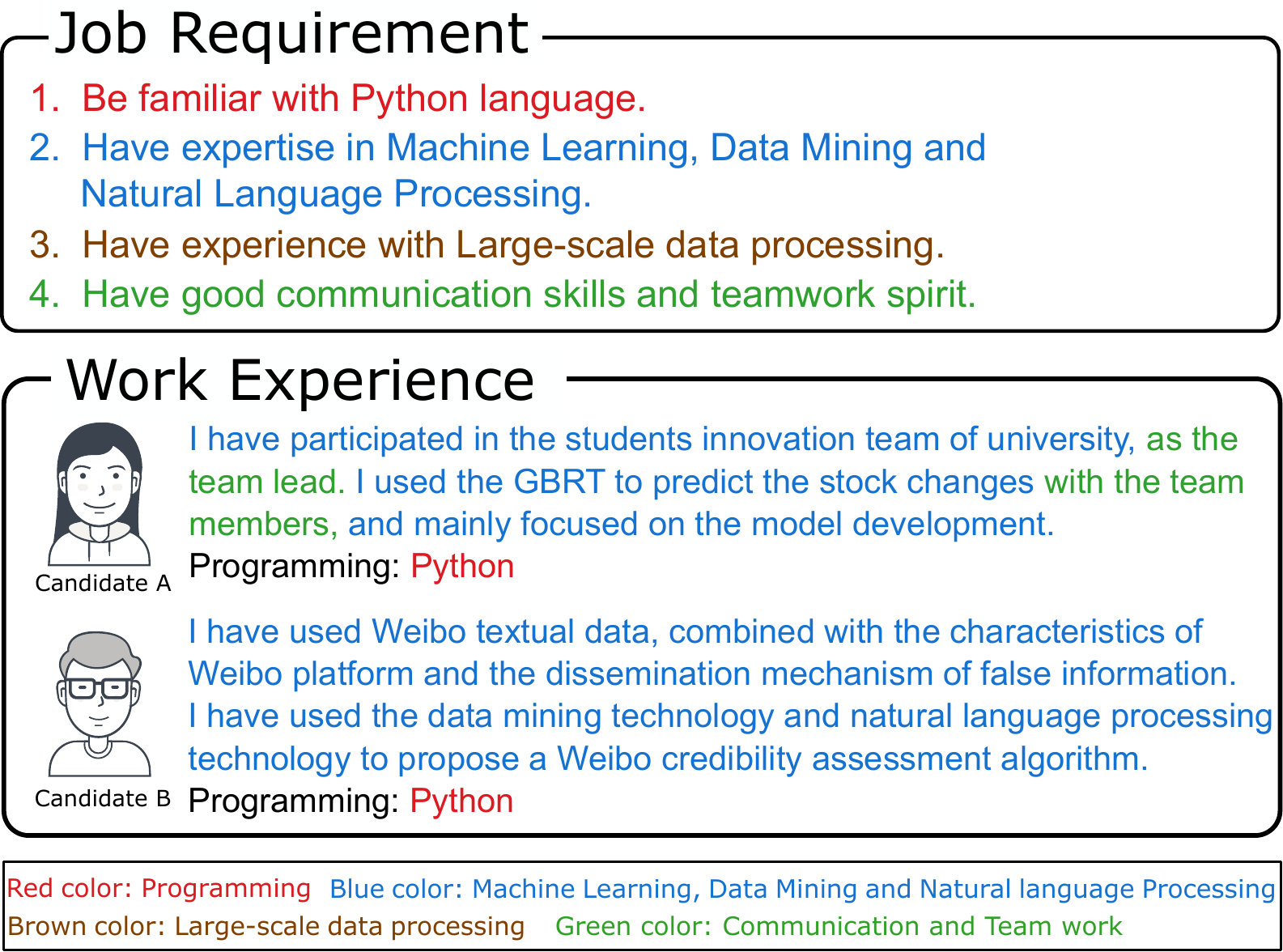}
  \caption{A motivating example of Person-Job Fit.}
  \label{job_resume_content}
  \vspace{-5mm}
\end{figure}

Indeed, as a crucial task for job recruitment, Person-Job Fit has been well studied from different perspectives, such as job-oriented skill measuring~\cite{xu2017measuring}, candidate matching~\cite{malinowski2006matching} and job recommendations~\cite{lee2007fighting,paparrizos2011machine,zhang2014research}. Along this line, some related tasks, such as talent sourcing~\cite{xu2016talent,zhu2016recruitment} and job transition~\cite{wang2013time} have also been studied. However, these efforts largely depend on the manual inspection of features or key phrases from domain experts, and thus lead to high cost and the inefficient, inaccurate, and subjective judgments.

To this end, in this paper, we propose an end-to-end Ability-aware Person-Job Fit Neural Network (APJFNN)
model, which has a goal of reducing the dependence on human labeling data and can provide better interpretation about the fitting results. The key idea of our approach is motivated by the example shown in Figure~\ref{job_resume_content}.  There are 4 requirements including 3 technical skill (\emph{programming, machine learning} and \emph{big data processing}) requirements and 1 comprehensive quality (\emph{communication and team work}) requirement. Since multiple abilities may fit the same requirement and different candidates may have different abilities, all the abilities should be weighed for a comprehensive score in order to compare the matching degree among different candidates. During this process, traditional methods, which simply rely on keywords/feature matching, may either ignore some abilities of candidates, or mislead recruiters by subjective and incomplete weighing of abilities/experiences from domain experts. Therefore, for developing more effective and comprehensive Person-Job Fit solution, abilities should be not only represented via the semantic understanding of rich textual information from large amount of job application data, but also automatically weighed based on the historical recruitment results.

Along this line, all the job postings and resumes should be comprehensively analyzed without relying on human judgement. To be specific, for representing both the job-oriented \textbf{abilities} and experiences of candidates, we first propose a word-level semantic representation based on Recurrent Neural Network (RNN) to learn the latent features of each word in a joint semantic space. Then, two hierarchical \textbf{ability-aware} structures are designed to guide the learning of semantic representation for job requirements as well as the  corresponding experiences of candidates. In addition, for measuring the importance of different abilities, as well as the relevance between requirements and experiences, we also design four hierarchical ability-aware attention strategies to highlight those crucial abilities or experience. This scheme will not only improve the performance, but also enhance the interpretability of matching results. Finally, extensive experiments on a large-scale real-world data set clearly validate the effectiveness of our APJFNN framework compared with several baselines.

\textbf{Overview.} The rest of this paper is organized as follows. In Section 2, we briefly introduce some related works of our study. In Section 3, we introduce the preliminaries and formally define the problem of Person-Job Fit. Then, technical details of our Ability-aware Person-Job Fit Neural Network will be introduced in Section 4. Afterwards, we comprehensively evaluate the model performance in Section 5, with some further discussions on the interpretability of results. Finally, in Section 6, we conclude the paper.

\begin{figure*}[t!]
  \centering
  \vspace{-5mm}
  \includegraphics[width=0.87\linewidth]{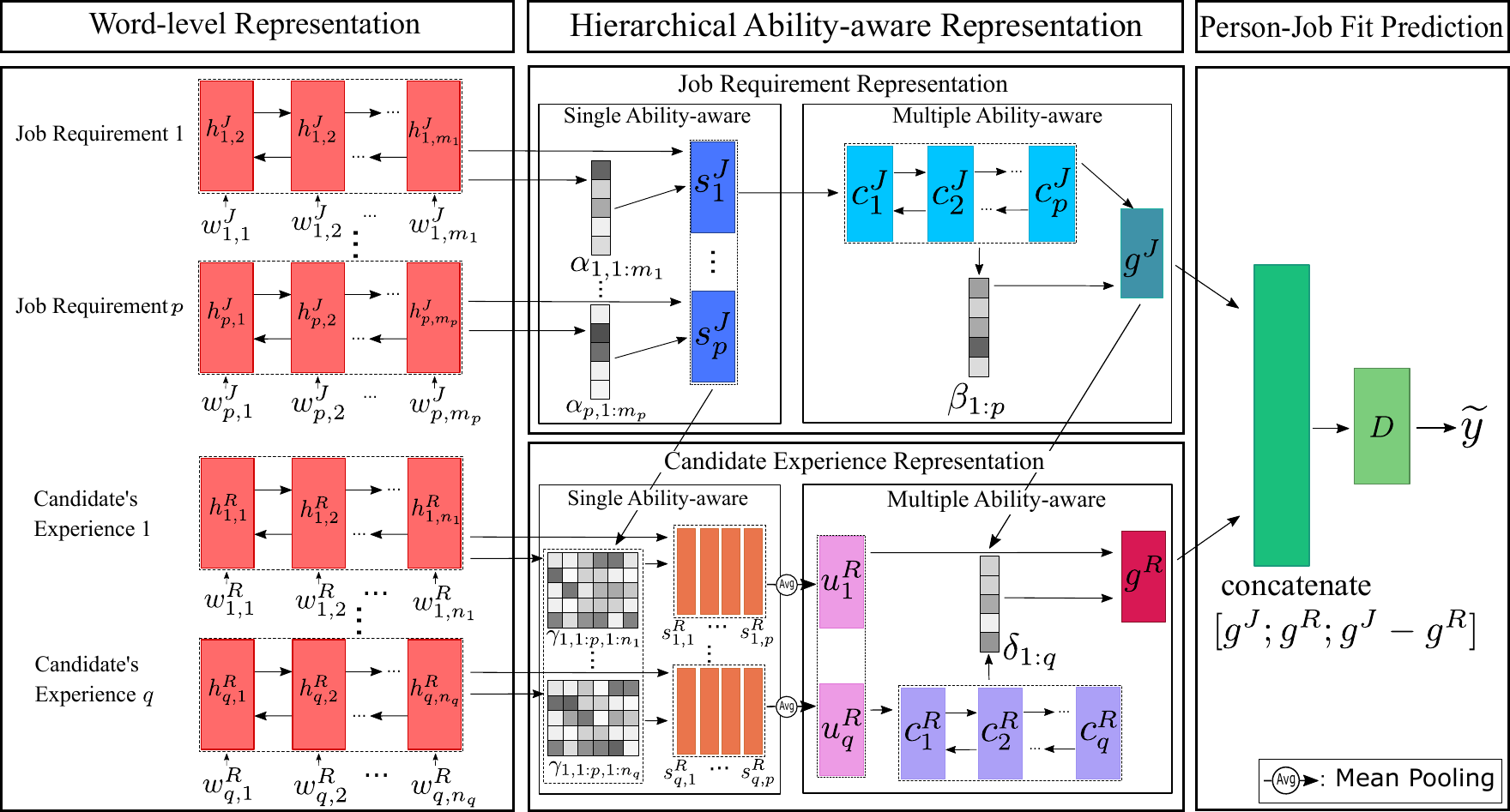}
  \caption{An illustration of the proposed Ability-aware Person-Job Fit Neural Network (APJFNN), which can be separated into three components, namely Word-level Representation, Hierarchical Ability-aware Representation and Person-Job Fit Prediction. Meanwhile, two different hierarchical structures are used to learn the ability-aware representation of job requirement and candidate experience respectively.}
\vspace{-2mm}
  \label{APJFNN}
  \vspace{-2mm}
\end{figure*}
\section{Related Work}
The related works of our study can be grouped into two categories, namely \emph{Recruitment Analysis} and \emph{Text Mining with Deep Learning}.

\subsection{Recruitment Analysis}

Recruitment is always a core function of human resource management to support the success of organizations. Recently, the newly available recruitment big data enables researchers to conduct recruitment analysis through more quantitative ways~\cite{zhu2016recruitment,harris2017finding,xu2017measuring,javed2017large,Lin-AAAI-17,xu2016talent}. In particular, the study of measuring the matching degree between the talent qualification and the job requirements, namely Person-Job Fit~\cite{sekiguchi2004person}, has become one of the most striking topics.

The early research efforts of Person-Job Fit can be dated back to \cite{malinowski2006matching}, where \textsl{Malinowski et al.} built a bilateral person-job recommendation system using the profile information from both candidates and jobs, in order to find a good match between talents and jobs. Then, \textsl{Lee~et~al.} followed the ideas of recommender systems and proposed a comprehensive job recommender system for job seekers, which is based on a broad range of job preferences and interests~\cite{lee2007fighting}. In~\cite{zhang2014research}, \textsl{Zhang et al.} compared a number of user-based collaborative filtering and item-based collaborative filtering algorithms on recommending suitable jobs for job seekers.

Recently, the emergence of various online recruitment services provides a novel perspective for recruitment analysis. For example, in~\cite{zhang2016glmix}, \textsl{Zhang et al.} proposed a generalized linear mixed models (GLMix), a more fine-grained model at the user or item level, in the LinkedIn job recommender system, and generated 20\% to 40\% more job applications for job seekers. In \cite{cheng2013jobminer}, \textsl{Cheng et al.} collected the job-related information from various social media sources and constructed an inter-company job-hopping network to demonstrate the flow of talents.  In \cite{wang2013time}, \textsl{Wang et al.} predicted the job transition of employees by exploiting their career path data. \textsl{Xu et al.} proposed a talent circle detection model based on a job transition network which can help the organizations to find the right talents and deliver career suggestions for job seekers to locate suitable jobs~\cite{xu2016talent}.

\subsection{Text Mining With Deep Learning}
Generally, the study of Person-Job Fit based on textual information can be grouped into the tasks of text mining, which is highly related to Natural Language Processing (NLP) technologies, such as text classification \cite{yang1997comparative,kim2014convolutional}, text similarity \cite{gomaa2013survey,severyn2015learning,kim2014convolutional}, and reading comprehension \cite{berant2014modeling,hermann2015teaching}. Recently, due to the advanced performance and flexibility of deep learning, more and more researchers try to leverage deep learning to solve the text mining problems. Compared with traditional methods that largely depend on the effective human-designed representations and input features~(e.g., word n-gram \cite{wang2012baselines}, parse trees \cite{cherry2008discriminative} and lexical features \cite{melville2009sentiment}), the deep learning based approaches can learn effective models for large-scale textural data without labor-intensive feature engineering.


Among various deep learning models, \textsl{Convolutional Neural Network} (CNN) \cite{lecun1998gradient} and \textsl{Recurrent Neural Network} (RNN) \cite{elman1990finding} are two representative and widely-used architectures, which can provide effective ways for NLP problems from different perspectives.

Specifically, CNN is efficient to extract local semantics and hierarchical relationships in textural data. For instance, as one of the representative works in this field, \textsl{Kalchbrenner et al.}~\cite{kalchbrenner2014convolutional} proposed a Dynamic Convolutional Neural Network (DCNN) for modeling sentences, which obtained remarkable performance in several text classification tasks. Furthermore, \textsl{Kim et al.} have shown that the power of CNN on a wide range of NLP tasks, even only using a single convolutional layer \cite{kim2014convolutional}. From then on, CNN-based approaches have attracted much more attentions on many NLP tasks. For example, in \cite{he2015multi}, \textsl{He et al.} used CNN to extract semantic features from multiple levels of granularity for measuring the sentences similarity. \textsl{Dong et al.} introduced a multi-column CNN for addressing the Question Answering problem~\cite{dong2015question}.

Compared with CNNs, RNN-based models are more ``natural'' for modeling sequential textual data, especially for the tasks of modeling serialization information, and learning the long-span relations or global semantic representation. For example, in \cite{tang2015document}, \textsl{Tang et al.} handled the document level sentiment classification with gated RNN. \textsl{Zhang et al.} designed a novel deep RNN model to perform the keyphrase extraction task \cite{zhang2016keyphrase}. Meanwhile, RNN also shows its effectiveness on several text generation tasks with the Encoder-Decoder framework. For example, in \cite{cho2014learning}, \textsl{Cho et al.} firstly used the framework for Machine Translation. \textsl{Bahdanau et al.} introduced an extension to this framework with the attention mechanism \cite{bahdanau2014neural} and validated the advantages of their model in translating long sentences. Similarly, in \cite{nallapati2016abstractive}, \textsl{Nallapati et al.} adapted the framework for automatic text summarization.


In this paper, we follow some outstanding ideas in the above works according to the properties of Person-Job Fit. And we propose an interpretable end-to-end neural model APJFNN based on RNN with four ability-aware attention mechanisms. Therefore, APJFNN can not only improve the performance of Person-Job Fit, but also enhance the model interpretability in practical scenarios.

\vspace{-2mm}
\section{Problem Formulation}
In this paper, we target at dealing with the problem of Person-Job Fit, which focuses on measuring the matching degree between job requirements in a \emph{job posting}, and the experiences in a \emph{resume}.

Specifically, to formulate the problem of Person-Job Fit, we use $J$ to denote a \textbf{job posting}, which contains $p$ pieces of \textbf{ability requirements}, denoted as $J=\{{j_1},{j_2},...,{j_{p}}\}$. For instance, there exist 4 requirements in Figure~\ref{job_resume_content}, thus $p=4$ in this case. Generally, we consider two types of ability requirements, i.e., the \textbf{professional skill} requirements (e.g., \emph{Data Mining} and \emph{Natural Language Processing} skills), and \textbf{comprehensive quality} requirements (e.g., \emph{Team Work}, \emph{Communication Skill} and \emph{Sincerity}). All the requirements will be analyzed comprehensively without special distinction by different types. Moreover, each $j_{l}$ is assumed to contain $m_l$ words, i.e., $j_l=\{j_{l,1},j_{l,2},...,j_{l,m_l}\}$.

Similarly, we use $R$ to represent a \textbf{resume} of a candidate, which includes $q$ pieces of \textbf{experiences}, i.e., $\{{r_1},{r_2},...,{r_{q}}\}$. In particular, due to the limitation of our real-world data, in this paper we mainly focus on the working experiences of candidate, as well as description of some other achievements, e.g., \emph{project experiences, competition awards} or \emph{research paper publications}. Besides, each experience $r_l$ is described by $n_l$ words, i.e., ${r_l}=\{r_{l,1},r_{l,2},...,r_{l,n_l}\}$.

Finally, we use $S$ to indicate a \emph{job application}, i.e., a Person-Job pair. Correspondingly, we have a \emph{recruitment result label} $y \in \{0, 1\}$ to indicate whether the candidate has passed the interview process, i.e., $y=1$ means a successful application, while $y=0$ means a failed one. What should be noted is that, one candidate is allowed to apply several jobs simultaneously, and one job position could be applied by multiple candidates. Thus, the same $J$ may exist in different $S$, so does $R$. Along this line, we can formally define the problem of Person-Job Fit as follow:
\begin{definition}
\textsl{(PROBLEM DEFINITION).} Given a set of job applications $\mathcal{S}$, where each application $S \in \mathcal{S}$ contains a job posting $J$ and a resume $R$, as well as the recruitment result label $y$. The target of Person-Job Fit is to learn a predictive model $M$ for measuring the matching degree between $J$ and $R$, and then corresponding result label $y$ could be predicted.
\end{definition}
In the following section, we will introduce the technical details of our APJFNN model for addressing the above problem.


\section{Ability-aware Person-Job Fit Neural Network}
As shown in Figure~\ref{APJFNN}, APJFNN mainly consists of three components, namely \textsl{Word-level Representation}, \textsl{Hierarchical Ability-aware Representation} and \textsl{Person-Job Fit Prediction}.

Specifically, in Word-level Representation, we first leverage an RNN to project words of job postings and resumes onto latent representations respectively, along with sequential dependence between words. Then, we feed the word-level representations into Hierarchical Ability-aware Representation, and extract the ability-aware representations for job postings and resumes simultaneously by hierarchical representation structures. To capture the semantic relationships between job postings and resumes and enhance the interpretability of model, we design four attention mechanisms from the perspective of ability to polish their representations at different levels in this component. Finally, the jointly learned representations of job postings and resumes are fed into Person-Job Fit Prediction to evaluate the matching degree between them.

\vspace{-1mm}
\subsection{Word-level Representation}
\begin{figure}[t]
\centering
\subfigure[]{
\includegraphics[width=1.5in]{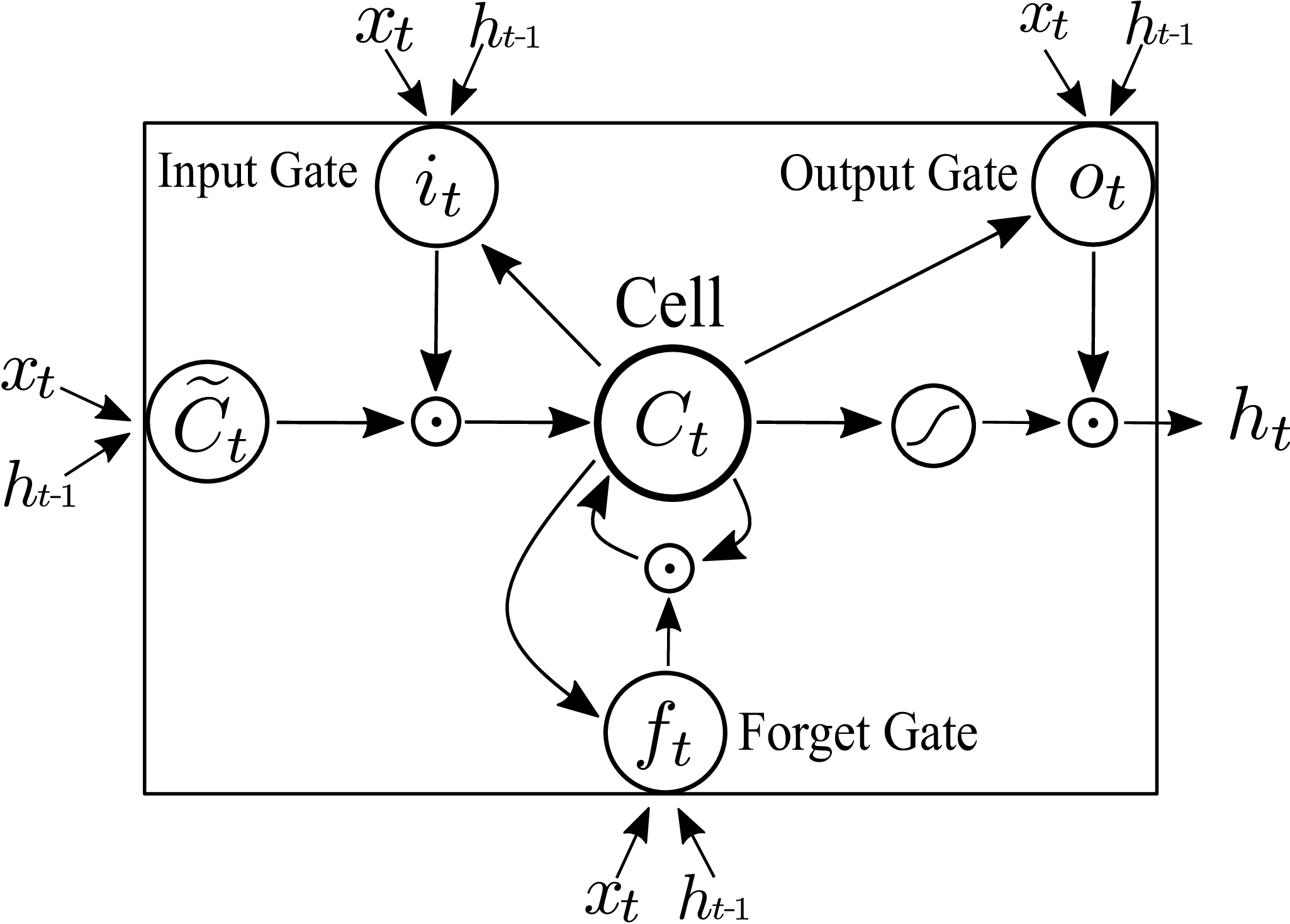}
\label{LSTM}}
\hspace{-0.06in}
\subfigure[]{
\includegraphics[width=1.5in]{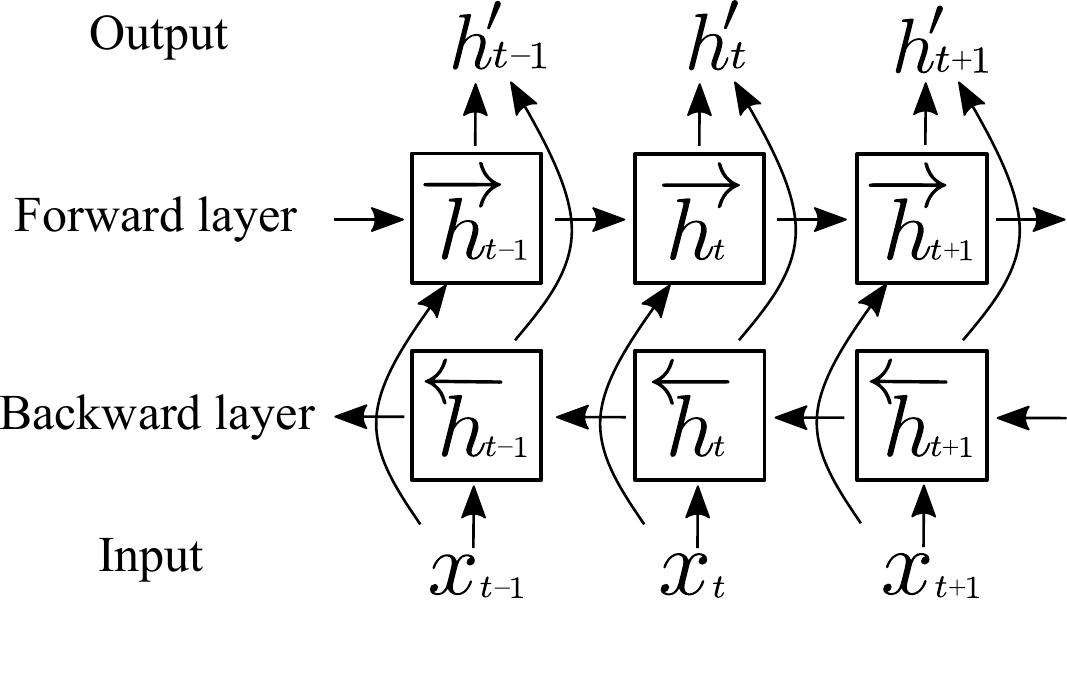}
\label{BiRNN}}
\vspace{-0.1in}
\caption{(a): The architecture of Long Short-Term Memory block with one cell. (b): The architecture of bidirectional recurrent neural network.}
\end{figure}

To embed the sequential dependence between words into corresponding representations, we leverage a special RNN, namely Bi-directional Long Short Term Memory network (BiLSTM), on a shared word embedding to generate the word-level representations for job postings and resumes. Compared with the vanilla RNN, LSTM \cite{hochreiter1997long} cannot only store and access a longer range of contextual information in the sequential input, but also handle the vanishing gradient problem in the meanwhile.
Figure~\ref{LSTM} illustrates a single cell in LSTM, which has a cell state and three gates, i.e., input gate $i$, forget gate $f$ and output gate $o$. Formally, the LSTM can be formulated as follows:
\begin{equation*}
\begin{split}
i_t &= \sigma(W_i[x_t,h_{t-1}]+b_i),\ f_t = \sigma(W_f[x_t,h_{t-1}]+b_f), \\
\widetilde{C}_t &= tanh(W_C[x_t,h_{t-1}]+b_C),\ C_t = f_t \odot C_{t-1}+i_t \odot\widetilde{C}_t, \\
o_t &= \sigma(W_o[x_t,h_{t-1}]+b_o),\ h_t = o_t \odot tanh(C_t),
\end{split}
\end{equation*}
where $X=\{x_1,x_2,...,x_m\}$ and $m$ denote the input vector and the length of $X$ respectively. And $W_f$, $W_i$, $W_C$, $W_o$, $b_f$, $b_i$, $b_C$, $b_o$ are the parameters as weight matrices and biases, $\odot$ represents element-wise multiplication, $\sigma$ is the sigmoid function, and $\{h_1,h_2,...,h_m\}$ represents a sequence of semantic features. Furthermore, the above formulas can be represented in short as:
\begin{equation*}
h_t=LSTM(x_t,h_{t-1}).
\end{equation*}
As shown in Figure~\ref{BiRNN}, the BiLSTM uses the input sequential data and their reverse to train the semantic vectors $\{h'_1,h'_2,...,h'_m\}$. The hidden vector $h'_t$ is the concatenation of the forward hidden vector $\overrightarrow{h_t}$ and backward hidden vector $\overleftarrow{h_t}$ at $t$-step. Specifically, we have
\begin{equation*}
\begin{split}
\overrightarrow{h_t} &= LSTM(x_t,\overrightarrow{h_{t-1}}), \\
\overleftarrow{h_t} &= LSTM({x_t},\overleftarrow{h_{t+1}}), \\
h'_t &= \left[ \overrightarrow{h_t};  \overleftarrow{h_t}  \right].
\end{split}
\end{equation*}
We can represent the above formulas in short as:
\begin{equation*}
h'_t=BiLSTM(x_{1:m},t), \ \ \forall t \in [1,...,m],
\end{equation*}
where $x_{1:m}$ denotes the input sequence $\{x_1,...,x_m\}$.

Now, we can use BiLSTM to model word-level representation in job posting $J$ and resume $R$. For $l$-th job requirement $j_l=\{j_{l,1},...,j_{l,m_l}\}$, we first embed the words to vectors by
\begin{equation*}
w^J_{l,t} = W_ej_{l,t}, \ \ w^J_{l,t} \in \mathbb{R}^{d_0},
\end{equation*}
where $w^J_{l,t}$ denotes $d_0$-dimensional word embedding of $t$-th word in ${j_l}$. As for $R$, word embedding $w^R_{l',t'}$ of $t'$-th word in candidate experience~$r_{l'}$ is generated by a similar way. It should be noted that the job postings and resumes share a same matrix $W_e$ which is initialized by a pre-trained word vector matrix and re-trained during training processing.


Then, for each word in the $l$-th job requirement $j_l$ and $l'$-th candidate experience $r_{l'}$, we can calculate the word-level representation $\{h^J_{l,1},h^J_{l,2},...,h^J_{l,m_l}\}$ and $\{h^R_{l',1},h^R_{l',2},...,h^R_{l',n_{l'}}\}$ by:
\begin{equation}
\label{word_level_representation}
\begin{split}
h^J_{l,t} &= BiLSTM(w^J_{l,1:m_l},t), \ \ \forall t \in [1,...,m_l],\\
h^R_{l',t'} &= BiLSTM(w^R_{l',1:n_{l'}},t'), \ \ \forall t' \in [1,...,n_{l'}],\\
\end{split}
\end{equation}
where $w^J_{l,1:m_l}$ and $w^R_{l',1:n_{l'}}$ denote the word vectors input sequences of $j_l$ and $r_{l'}$, respectively. And $h^J_{l,t}$,$h^R_{l',t'}$ are $d_0$-dimension semantic representations of the $t$-th word in the $l$-th job requirement $j_l$ and $t'$-th word in the $l'$-th candidate experience $r_{l'}$.

\subsection{Hierarchical Ability-Aware Representation}
After getting the representations of job postings and resumes at word-level, we further extract more high-level representations for them. As for job postings, we consider that each ability requirement refers to a specific need of a job, and the entire needs of a job can further be summarized from all of its requirements. Following this intuition, we design a hierarchical neural network structure to model such hierarchical representation. And as for resumes, similar hierarchical relationships also exist between a candidate experiences and her qualification, thus a similar hierarchical neural network structure is also applied for resumes.

Besides, as we know, both of job postings and resumes are documents with relatively well-defined formats. For example, most of candidates tend to separate their past experiences by work contents and order them by time for facilitating understanding. Indeed, such kinds of format can help us to better extract representations. Thus, to improve the performance and interpretability, we follow the above intuitions and design four attention mechanisms to polish representations extracted by our model at different levels.

Specifically, this component can further be divided into four parts: 1) \textsl{Single Ability-aware in Job Requirement} for getting the semantic representation of each requirement in a job posting; 2) \textsl{Multiple Ability-aware in Job Requirement} for further extracting entire representation of a job posting, 3) \textsl{Single Ability-aware in Candidate Experience} for highlighting some experiences in resumes by ability requirements; 4) \textsl{Multiple Ability-aware in Candidate Experience} for finally profiling candidates with all previous experiences. In the following, we will introduce the technical details of each component.

\vspace{1mm}

\noindent\textbf{$\bullet$ Single Ability-aware in Job Requirement.} It is obvious that the meaning of a sentence is dominated by several keywords or phrases. Thus, to better capture the key information for each ability requirement, we use an attention mechanism to estimate the importance of each word in it.

This attention layer is the weighted sum of the semantic vector of each word in each ability requirement. Specifically, for $l$-th ability requirement ${j_l}$, we first use the word representation $\{h^J_{l,t},...,h^J_{l,m_l}\}$ as input of a fully-connected layer and calculate the similarity with word level context vector. Then, we use a softmax function to calculate the attention score $\alpha$, i.e.,
\begin{equation*}
\begin{split}
\alpha_{l,t} &= \frac{exp(e^J_{l,t})}{\sum_{z=1}^{m_l}exp(e^J_{l,z})},\\
e^J_{l,t} &= {v_{\alpha}}^{\rm T}tanh(W_{\alpha} h^J_{l,t}+b_{\alpha}), \\
\end{split}
\end{equation*}
where $v_{\alpha}$, $W_{\alpha}$ and $b_{\alpha}$ are the parameters to be learned during the training processing. Specifically, $v_{\alpha}$ denotes the context vector of the ${j_l}$, which is randomly initialized. The attention score $\alpha$ can be seen as the importance of each word in $j_l$.
Finally, we calculate the single ability-aware requirement representation $s^J_l$ for ${j_l}$ by:
\begin{equation}
\label{single_ability_representation}
s^J_l = \sum_{t=1}^{m_l}\alpha_{l,t}h^J_{l,t}.
\end{equation}

\vspace{1mm}

\noindent\textbf{$\bullet$ Multiple Ability-aware in Job Requirement.} In this part, we leverage the representations extracted by Single Ability-aware in Job Requirement to summarize the general needs of jobs. In most of jobs, although different ability requirements refer to different specific needs, their importance varies a lot. For example, for recruiting a software engineer, education background is much less important than professional skills.
Moreover, the order of ability requirements in job description will also reflect their importance. With these intuitions, we first use a BiLSTM to model the sequential information of ability requirements. Then we add an attention layer to learn the importance of each ability requirement. Formally, sequential ability representation~$\{s^J_1,...,{s^J_p}\}$, learned in Single Ability-aware in Job Requirement, are used as input of a BiLSTM to generate a sequence of hidden state vectors $\{c^J_1,...,c^J_p\}$, i.e.,
\begin{equation*}
c^J_t = BiLSTM(s^J_{1:p},t), \ \ \forall t \in [1,...,p].
\end{equation*}

Similar with the first attention layer, we add another attention layer above the LSTMs to learn importance of each ability requirement. Specifically, we calculate the importance $\beta_t$ of each ability requirement $j_t$ based on the similarity between its hidden state $c^J_t$ and the context vector $v_{\beta}$ of all the ability requirements, i.e.,
\begin{equation*}
\begin{split}
\beta_{t} &= \frac{exp(f^J_{t})}{\sum_{z=1}^{p}exp(f^J_{z})}, \\
f^J_{t} &= {v_{\beta}}^{\rm T}tanh({W_{\beta}}{c^{J}_t}+b_{\beta}), \\
\end{split}
\end{equation*}
where the parameters $W_{\beta}$, $b_{\beta}$ and context vector $v_{\beta}$ are learned during training. Then, a latent multiple ability-aware job requirement vector will be calculated by weighted sum of the hidden state vectors of abilities, i.e.,
\begin{equation*}
g^J = \sum_{t=1}^{p}\beta_{t}c^J_t.
\end{equation*}
Particularly, the attention scores $\beta$ can greatly improve the interpretability of the model. It is helpful for visualizing the importance of each ability requirement in practical recruitment applications.


\vspace{1mm}

\noindent\textbf{$\bullet$ Single Ability-aware in Experience.} Now we turn to introduce the learning of resume representations. Specifically, when a recruiter examines whether a candidate matches a job, she tends to focus on those specific skills related to this job, which can be reflected by the candidate experiences. As shown in Figure~\ref{job_resume_content}, for candidate A, considering the fourth job requirement, we will pay more attention to the highlighted ``green'' sentences. Meanwhile, we may focus on the ``blue'' sentences when matching the second requirement. Thus, we design a novel ability-aware attention mechanism to qualify the ability-aware contributions of each word in candidate experience to a specific ability requirement.
Formally, for the $l$-th candidate experience $r_l$, its word-level semantic representation is calculated by a BiLSTM. And we use an attention-based relation score $\gamma_{l,k,t}$ to qualify the ability-aware contribution of each semantic representation $h^R_{l,t}$ to the $k$-th ability requirement $j_k$. It can be calculated by
\begin{equation*}
\begin{split}
\gamma_{l,k,t} &= \frac{exp(e^R_{l,k,t})}{\sum_{z=1}^{n_l}exp(e^R_{l,k,z})},\\
e^R_{l,k,t} &= v_{\gamma}^{\rm T}tanh(W_{\gamma}s^J_k+U_{\gamma}h^R_{l,t}),
\end{split}
\end{equation*}
where the $W_{\gamma}$,$U_{\gamma}$, $v_{\gamma}$ are parameters, $s^J_k$ is the semantic vector of ability requirement $j_k$ which is calculated by Equation~\ref{single_ability_representation}.

Finally, the single ability-aware candidate experience representation is calculated by the weighted sum of the word-level semantic representation of $r_l$
\begin{equation*}
s^R_{l,k} = \sum_{t=1}^{n_l}\gamma_{l,k,t}h^R_{l,t} .
\end{equation*}

Here, the attention score $\gamma$ further enhances the interpretability of APJFNN. It enables us to understand whether and why a candidate is qualified for an ability requirement, we will further give a deep analysis in the experiments.

\vspace{1mm}

\noindent\textbf{$\bullet$ Multiple Ability-aware in Experience.} For a candidate, her ordered experiences can reveal her growth process well and such temporal information can also benefit the evaluation on her abilities.
To capture such temporal relationships between experiences, we leverage another BiLSTM. Specifically, we first add a mean-pooling layer above the single ability-aware candidate experience representation to generate the latent semantic vector $u^R_l$ for $l$-th candidate experience $r_l$.
\begin{equation*}
u^R_l = \frac{\sum_{t=1}^{p}s^R_{l,t}}{p} .
\end{equation*}

Now we get a set of semantic vectors for candidate experiences, that is $\{u^R_1,...,u^R_q\}$. Considering there exist temporal relationships among $\{u^R_1,...,u^R_q\}$, we use a BiLSTM to chain them, i.e.,
\begin{equation*}
c^R_t = BiLSTM(u^R_{1:q},t), \ \ \forall t \in [1,...,q].
\end{equation*}
Finally, we use the weighted sum of the hidden states $\{c^R_1,...,c^R_q\}$ to generate the multiple ability-aware candidate experience representation, i.e.,
\begin{equation*}
\begin{split}
\delta_{t} &= \frac{exp(f^R_{t})}{\sum_{z=1}^{q}exp(f^R_{z})},\\
f^R_{t} &= v_{\delta}^{\rm T}tanh(W_{\delta}g^J+U_{\delta}c^R_t), \\ 
g^R &= \sum_{t=1}^{q}\delta_{t}c^R_t. 
\end{split}
\end{equation*}

\subsection{Person-Job Fit Prediction}
With the process of Hierarchical Ability-aware Representation, we can jointly learn the representations for both job postings and resumes.
To measure the matching degree between them, we finally treat them as input and apply a comparison mechanism based on a fully-connected network to learn the overall Person-Job Fit representation $D$ for predicting the label $\widetilde{y}$ by a logistic function. The mathematical definition is as follows.
\begin{equation}
\label{D_eq}
\begin{split}
D &= tanh(W_d[g^J;g^R;g^J-g^R]+b_d), \\
\widetilde{y} &= {Sigmoid}(W_yD+b_y),
\end{split}
\end{equation}
where $W_d$,$b_d$,$W_y$,$b_y$ are the parameters to tune the network and $\widetilde{y} \in [0,1]$. Meanwhile, we minimize the binary cross entropy to train our model.

\section{Experiments}
In this section, we will introduce the experimental results based on a real-world recruitment data set. Meanwhile, some case studies are demonstrated for revealing interesting findings obtained by our model APJFNN.

\subsection{Data Description}
In this paper, we conducted our validation on a real-world data set, which was provided by a high tech company in China. To protect the privacy of candidates, all the job application records were anonymized by deleting personal information.

\begin{figure*}[t]\centering
\vspace{-5mm}
\subfigure[]{
\includegraphics[width=2.2in]{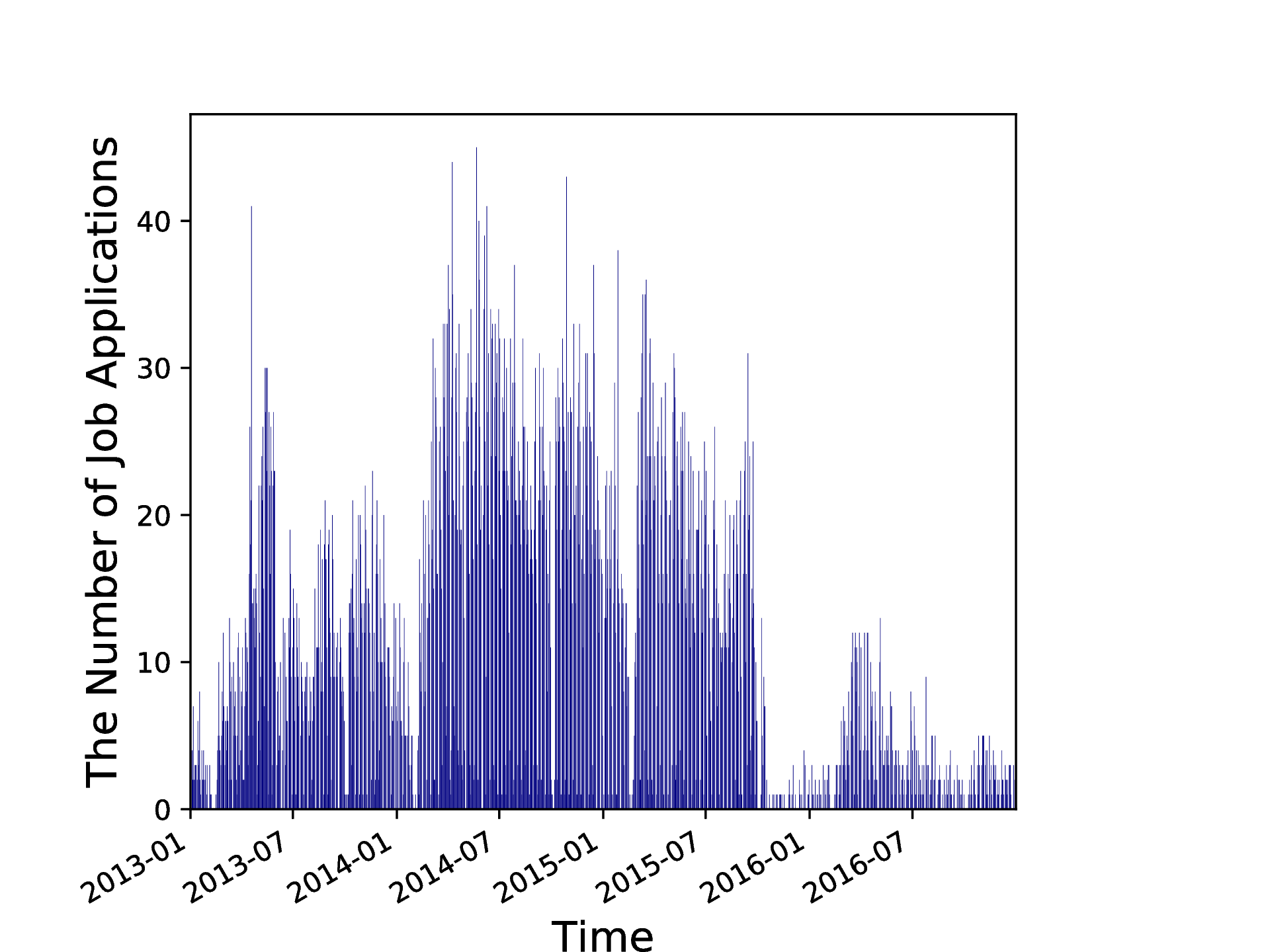}
\label{time_dis}}
\vspace{-3mm}
\hspace{-0.15in}
\subfigure[]{
\includegraphics[width=2.2in]{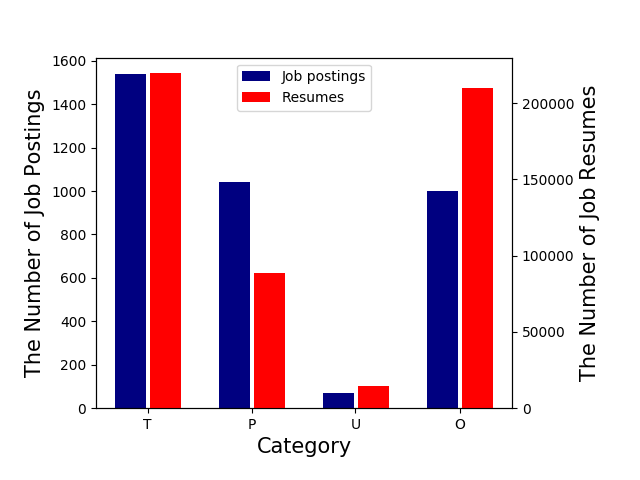}
\label{category}}
\hspace{-0.15in}
\subfigure[]{
\includegraphics[width=2.2in]{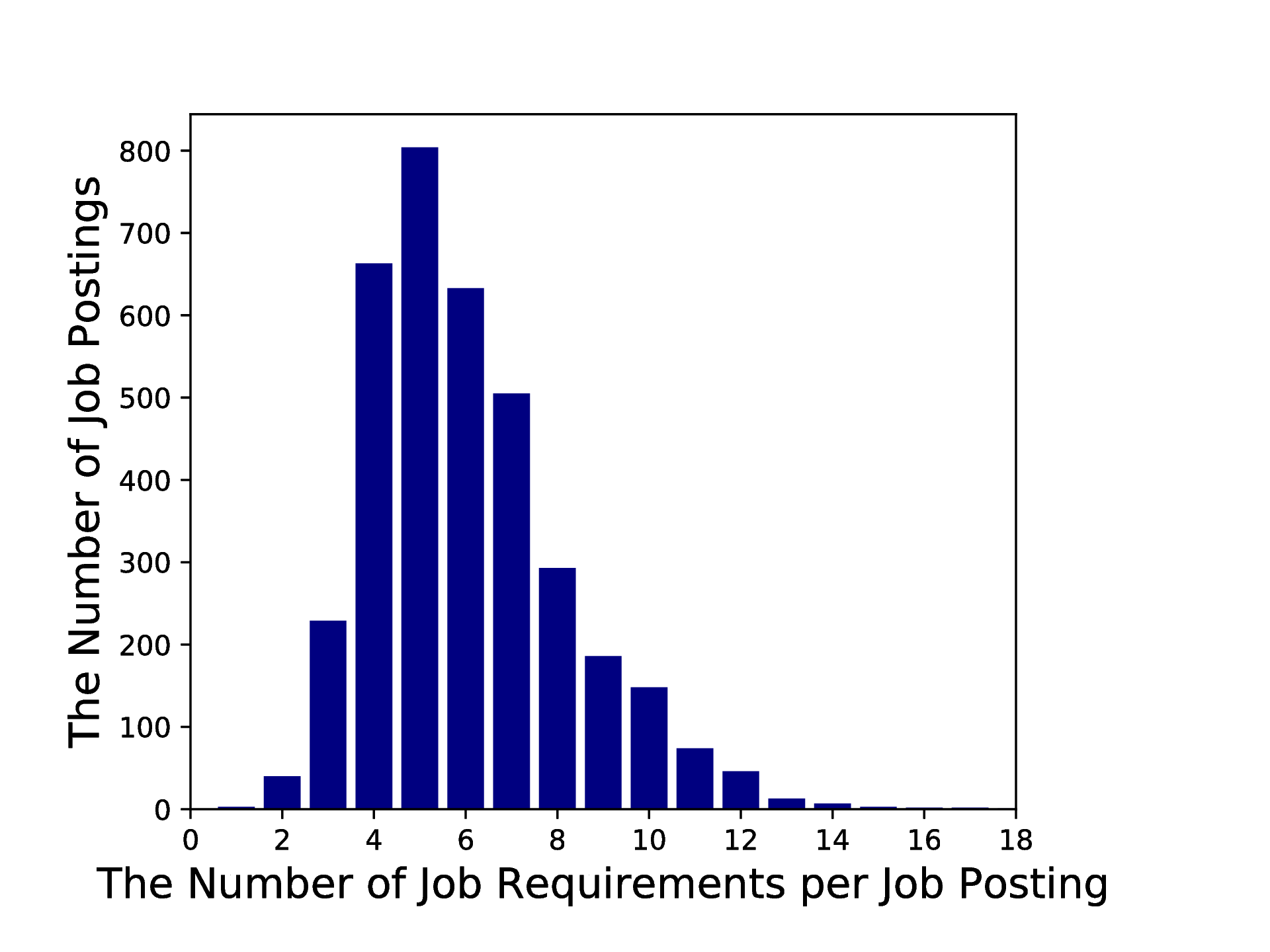}
\label{post_count}}
\vspace{-5mm}
\subfigure[]{
\includegraphics[width=2.2in]{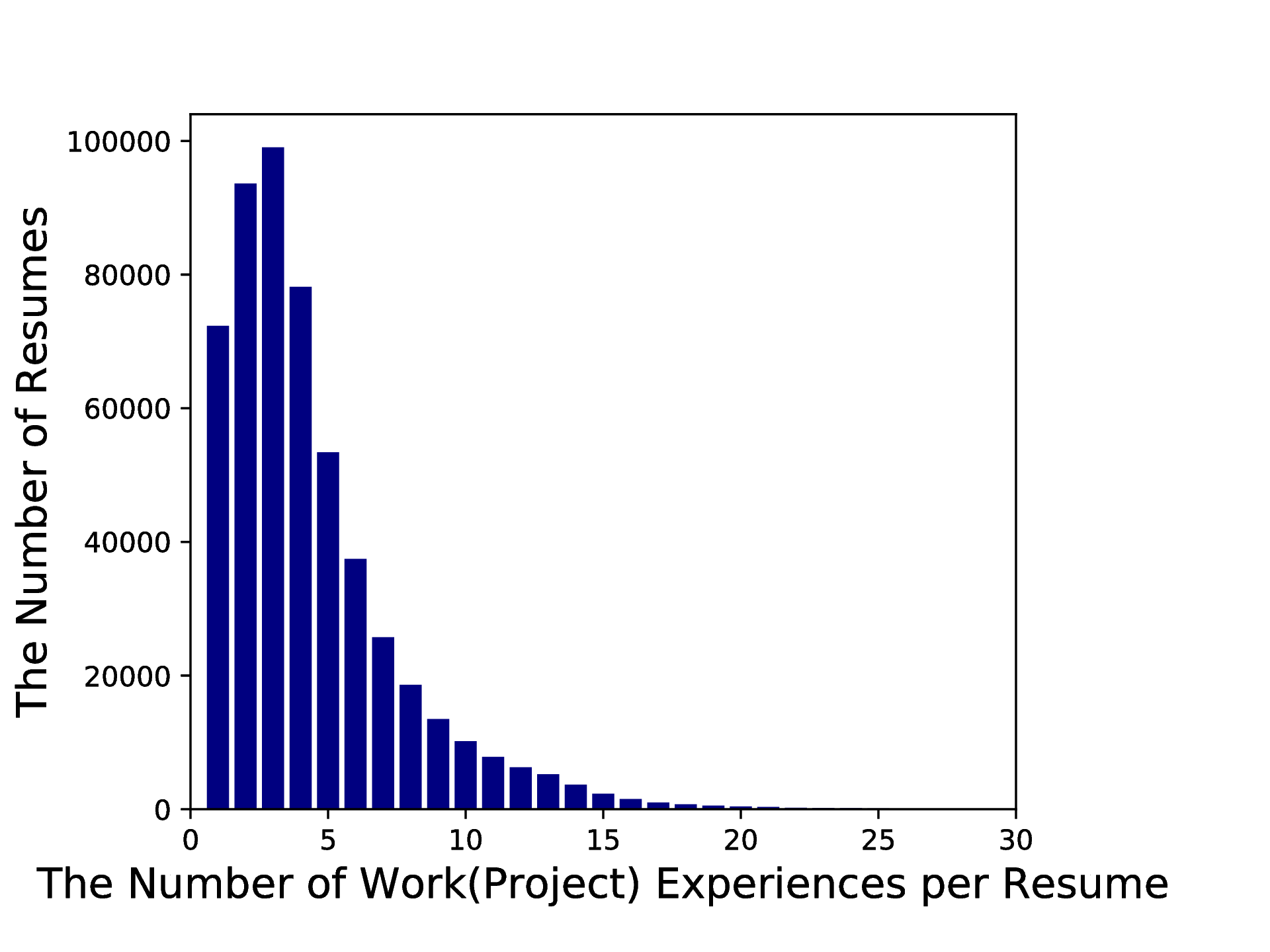}
\label{resume_count}}
\hspace{-0.15in}
\subfigure[]{
\includegraphics[width=2.2in]{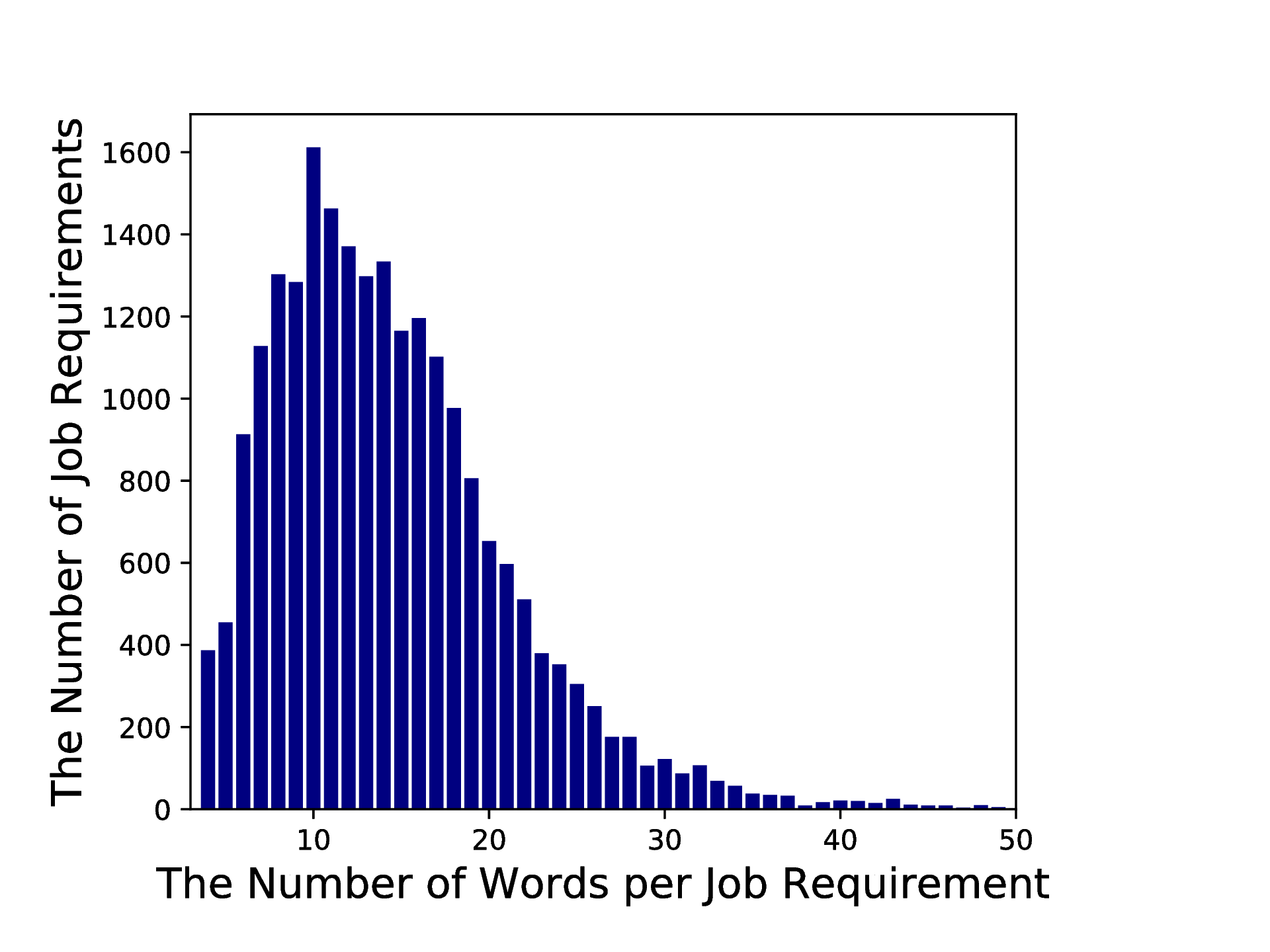}
\label{post_word}}
\hspace{-0.15in}
\subfigure[]{
\includegraphics[width=2.2in]{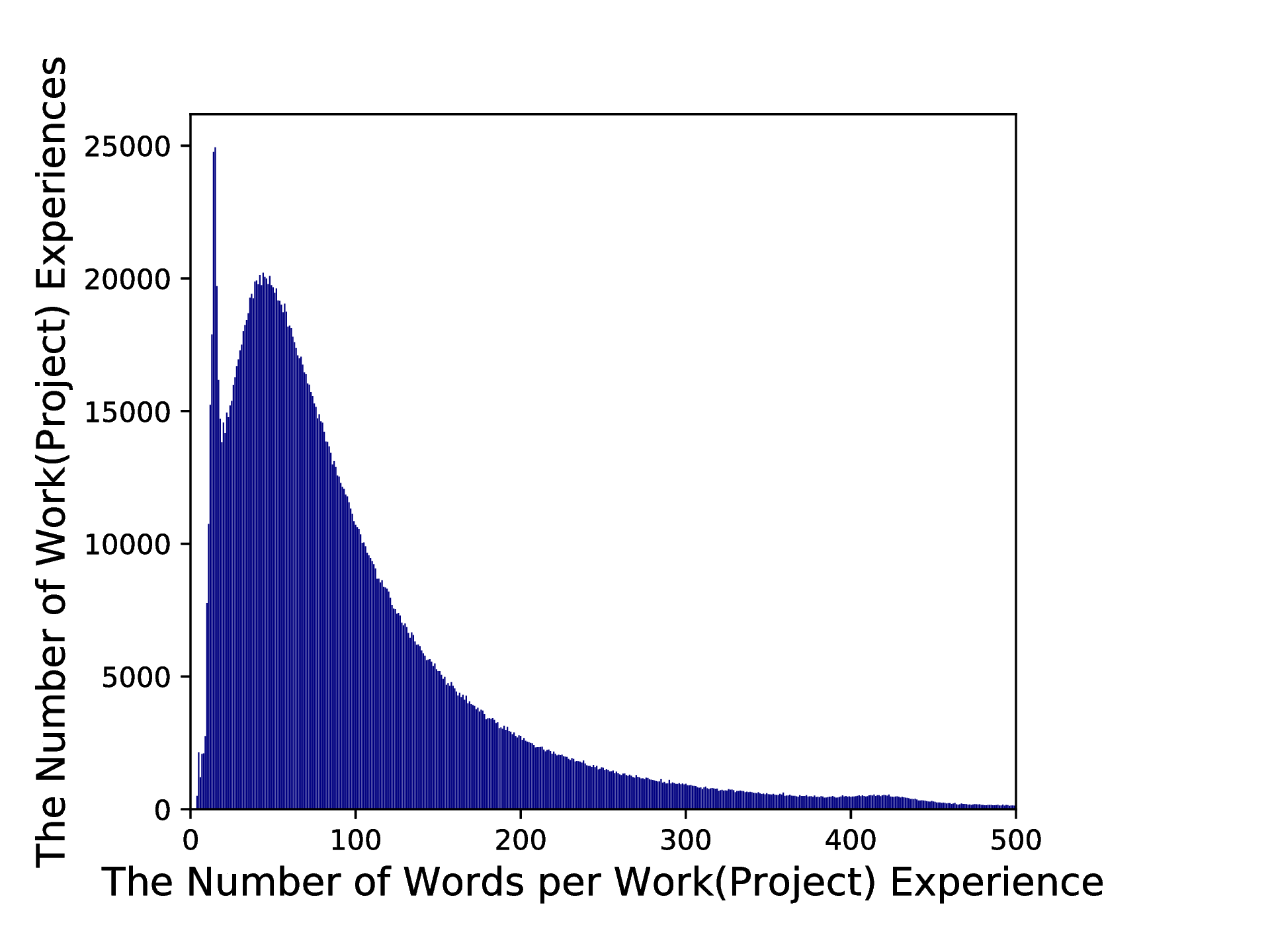}
\label{resume_word}}

\caption{(a): The time distribution of successful job applications. (b): The distribution of different categories w.r.t job posting and resume respectively. (c): The distribution of job requirements. (d): The words distribution of job requirement. (e): The distribution of candidate experiences. (f): The words distribution of candidate experience.}
\end{figure*}

The data set consists of \emph{17,766} job postings and \emph{898,914} resumes with a range of several years. Specifically, four categories of job postings, namely \emph{Technology}, \emph{Product}, \emph{User Interface} and \emph{Others} were collected. Figure~\ref{category} summarizes the distribution of job postings and resumes, according to different categories. We find that most of the applications are technology-oriented, and only about 1\% applications were accepted, which highlights the difficulty of talent recruitment. To a certain degree, this phenomenon may also explain the practical value of our work, as the results of Person-Job Fit may help both recruiters and job seekers to enhance the success rate.

Along this line, to ensure the quality of experiments, those incomplete resume (e.g., resumes without any experience records) were removed. Correspondingly, those job postings without any successful applications were also removed. Finally, \emph{3,652} job postings, \emph{12,796} successful applications and \emph{1,058,547} failed ones were kept in total, which lead to a typical imbalanced situation. Some basic statistics of the pruned data set are summarized in Table~\ref{statistics}. What should be noted is that, it is reasonable to have more applications than the number of resumes, since one candidate could apply several positions at the same time, which is mentioned above.
\begin{table}[t]
\caption{The statistics of the dataset}
\begin{tabular}{ l | r }
\hline\hline	
  Statistics & Values \\ \hline\hline
  \# of job postings &  3,652\\
  \# of resumes & 533,069  \\
  \# of successful applications & 12,796\\
  \# of failed applications & 1,058,547\\
  Average job requirements per posting & 6.002\\
  Average project/work experiences per resume & 4.042\\
  Average words per job requirement & 9.151\\
  Average words per project/work experience & 65.810 \\
  \hline\hline
\end{tabular}
\label{statistics}
\end{table}

\subsection{Experimental Setup}
Here, we introduce the detailed settings of our experiments, including the technique of word embedding, parameters for our APJFNN, as well as the details of training stage.

\vspace{1mm}
\noindent \textbf{$\bullet$ Word Embedding.}
First, we explain the embedding layer, which is used to transfer the original \emph{``bag of words''} input to a dense vector representation. In detail, we first used the Skip-gram Model~\cite{mikolov2013distributed} to pre-train the word embedding from job requirements and candidate's experiences. Then, we utilized the pre-trained word embedding results to initialize the embedding layer weight $W_e$, which was further fine-tuned during the training processing of APJFNN. Specifically, the dimension of word vectors was set to 100.

\vspace{1mm}
\noindent \textbf{$\bullet$ APJFNN Setting.}
In APJFNN model, according to the observation in Figure~\ref{post_count},~\ref{resume_count},~\ref{post_word} and~\ref{resume_word}, we set both the maximum number of job requirements in each job posting as 15, and so does the constraint of candidate experiences in each resume. Then, the maximum number of words in each requirement/experience was set as 30 and 300, respectively. Along this line, the excessive parts were removed. Also, the dimension of hidden state in BiLSTM was set as 200 to learn the word-level joint representation and requirement/experience representation. Finally, the dimension of parameters to calculate the attention score $\alpha$ and $\beta$ were set as 200, as well as 400 for $\gamma$ and $\delta$.

\vspace{1mm}
\noindent \textbf{$\bullet$ Training Setting.}
Following the idea in~\cite{glorot2010understanding}, we initialized all the matrix and vector parameters in our APJFNN model with uniform distribution in $[-\sqrt{6/(n_{in}+n_{out})},\sqrt{6/(n_{in}+n_{out})}]$, where $n_{in}$, $n_{out}$ denote the number of the input and output units, respectively. Also, models were optimized by using Adam~\cite{kingma2014adam} algorithm. Moreover, we set batch size as 64 for training, and further used the dropout layer with the probability 0.8 in order to prevent overfitting.

\begin{figure}
  \includegraphics[width=\linewidth]{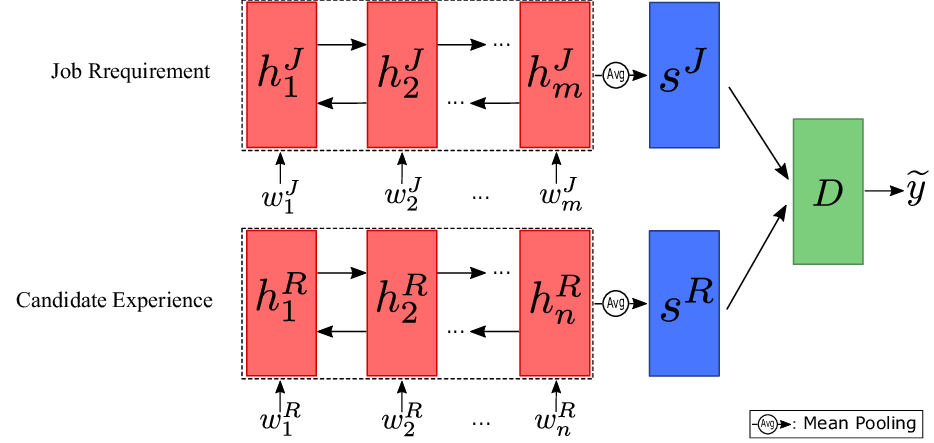}
  \caption{An illustration of the proposed Basic Person-Job Fit Neural Network (BPJFNN)}
  \label{BPJFNN}
  \vspace{-3mm}
\end{figure}

\subsection{Baseline Methods}
To validate the performance of our APJFNN model, several state-of-the-art supervised models were selected as baseline methods, including the classic supervise learning methods like~\textsl{Logistical Regression} (\textbf{LR}),~\textsl{Decision Tree} (\textbf{DT}),~\textsl{Adaboost} (\textbf{AB}),~\textsl{Random Forests} (\textbf{RF}) and~\textsl{Gradient Boosting Decision Tree} (\textbf{GBDT}). For these baselines, we used two kinds of input features to construct the experiment, separately.
\begin{itemize}
\item \textbf{Bag-of-words vectors.} We first created the bag-of-words vectors of ability requirements and candidate experiences respectively, where the $i$-th dimension of each vector is the frequency of the $i$-th word in dictionary. Then, two vectors were spliced together as input.

\item \textbf{Mean vector of word embedding.} We respectively averaged the pre-trained word vector of the requirements and experiences, and then spliced them as model input.
\end{itemize}

Besides, we also propose an RNN-based model called \textbf{B}asic \textbf{P}erson-\textbf{J}ob \textbf{F}it \textbf{N}eural \textbf{N}etwork (\textbf{BPJFNN}) as baseline, which could be treated as a simplified version of our APJFNN model. The structure of BPJFNN model is shown in Figure~\ref{BPJFNN}. To be specific, in this model, two BiLSTM are used to get the semantic representation of each word in requirements and experiences. What should be noted is that, here we treat all the ability requirements in one job posting as a unity, i.e., a ``long sentence'', instead of separate requirements, and so do the experiences in candidate resumes. Then, we add a mean-pooling layer above them to got two semantic vectors $s^J$, $s^R$, respectively. Finally, we can use following equations to estimate the Person-Job Fit result label $\widetilde{y}$.
\begin{equation*}
\begin{split}
D &= tanh(W_d\left[s^J;s^R;s^J-s^R\right]+b_d), \\
\widetilde{y} &= {softmax}(W_yD+b_y),
\end{split}
\end{equation*}
where the $W_d$ and $b_d$ are the parameters to learn.

\begin{table}[t]
\caption{The performance of APJFNN and baselines.}
\resizebox{\columnwidth}{!}{%
\begin{tabular}{ l | c | c | c | c | c }
\hline \hline
  Methods & Accuracy & Precision & Recall & F1 & AUC\\ \hline\hline 
  LR & 0.6228 &	 0.6232 & 	0.6261 & 	0.6246 & 0.6787 \\
  AB &	 0.6905 &	0.7028 & 0.6628 &	0.6822 &	 0.7642 \\
  DT & 0.6831 &	0.7492 &	 0.5527 &	0.6361 & 0.7355 \\
  RF & 0.7023 & 0.7257 & 0.6526 & 0.6872 &  0.7772 \\
  GBDT & 0.7281 & 0.7517 &	0.6831 &	0.7157 &	0.8108 \\ \hline
  LR (with word2vec) & 0.6479 & 0.6586 & 0.6175 & 0.6374 & 0.6946 \\
  AB (with word2vec) & 0.6342 &	0.6491 &	0.5878 &0.6170 & 0.6823 \\
  DT (with word2vec) & 0.5837 &	0.5893 &	0.5589 & 0.5737 &	0.6249 \\
  RF (with word2vec) & 0.6358 & 0.6551 &	0.5769 &	0.6135 &	0.7020 \\
  GBDT (with word2vec) & 0.6389 &	0.6444 &	0.6237 &	0.6339 & 0.7006 \\ \hline
  BPJFNN & 0.7156 &	0.7541 & 0.6417 & 0.6934 &	0.7818 \\
  \textbf{APJFNN} &  \textbf{0.7559} & \textbf{0.7545} & \textbf{0.7603} & \textbf{0.7574} & \textbf{0.8316} \\
  \hline\hline
\end{tabular}}
\label{real-world}
\vspace{-1mm}
\end{table}

\subsection{Evaluation Metrics}


Since, in the real-world process of talent recruitment, we usually have a potential \emph{``threshold''} to pick up those adequate candidate, which results in a certain \emph{``ratio of acceptance''}. However, we could hardly determine the acceptance rate properly, as it could be a personalized value which is affected by complicated factors. Thus, to comprehensively validate the performance, we selected the \textbf{AUC} index to measure the performance under different situations. Besides, we also adopted the \textbf{Accuracy}, \textbf{Precision}, \textbf{Recall} and \textbf{F1-measure} as the evaluation metrics.

\subsection{Experimental Result}
\vspace{1mm}
\noindent\textbf{$\bullet$ Overall Results.}
We conducted the task of Person-Job Fit based on the real-word data set, i.e., we used the successful job applications as \emph{positive samples}, and then used the failed applications as the \emph{negative instance} to train the models. In order to reduce the impact of imbalances in data, we used the under-sampling method to randomly select negative instances that are equal to the number of positive instances for each job posting to evaluate our model~\footnote{Since there were some job postings which the number of failed applications was less than the number of successful applications, we finally got 12,762 negative samples. The number of training, validation, testing samples is 20,446, 2,556 and 2,556 respectively.}. Along this line, we randomly selected 80\% of the data set as training data, another 10\% for tuning the parameters, and the last 10\% as test data to validate the performance.


The performance is shown in Table~\ref{real-world}. According to the results, clearly, we realize that our APJFNN outperforms all the baselines with a significant margin, which verifies that our framework could well distinguish those adequate candidates with given job postings. Especially, as APJFNN performs better than BPJFNN, it seems that our attention strategies could not only distinguish the critical ability/experience for better explanation, but also improve the performance with better estimation of matching results. 

At the same time, we find that almost all the baselines using the Bag-of-Words as input feature outperform those using the pre-trained word vector as input features (i.e., those with ``word2vec'' in Table~\ref{real-world}). This phenomenon may indicate that the pre-trained word vectors are not enough to characterize the semantic features of the recruitment textural data, this is the reason of why we use the BiLSTM above the embedding layer to extract the word-level semantic word representation.

\begin{figure}\centering
\vspace{-2mm}
\subfigure[The F1 performance]{
\includegraphics[width=1.7in]{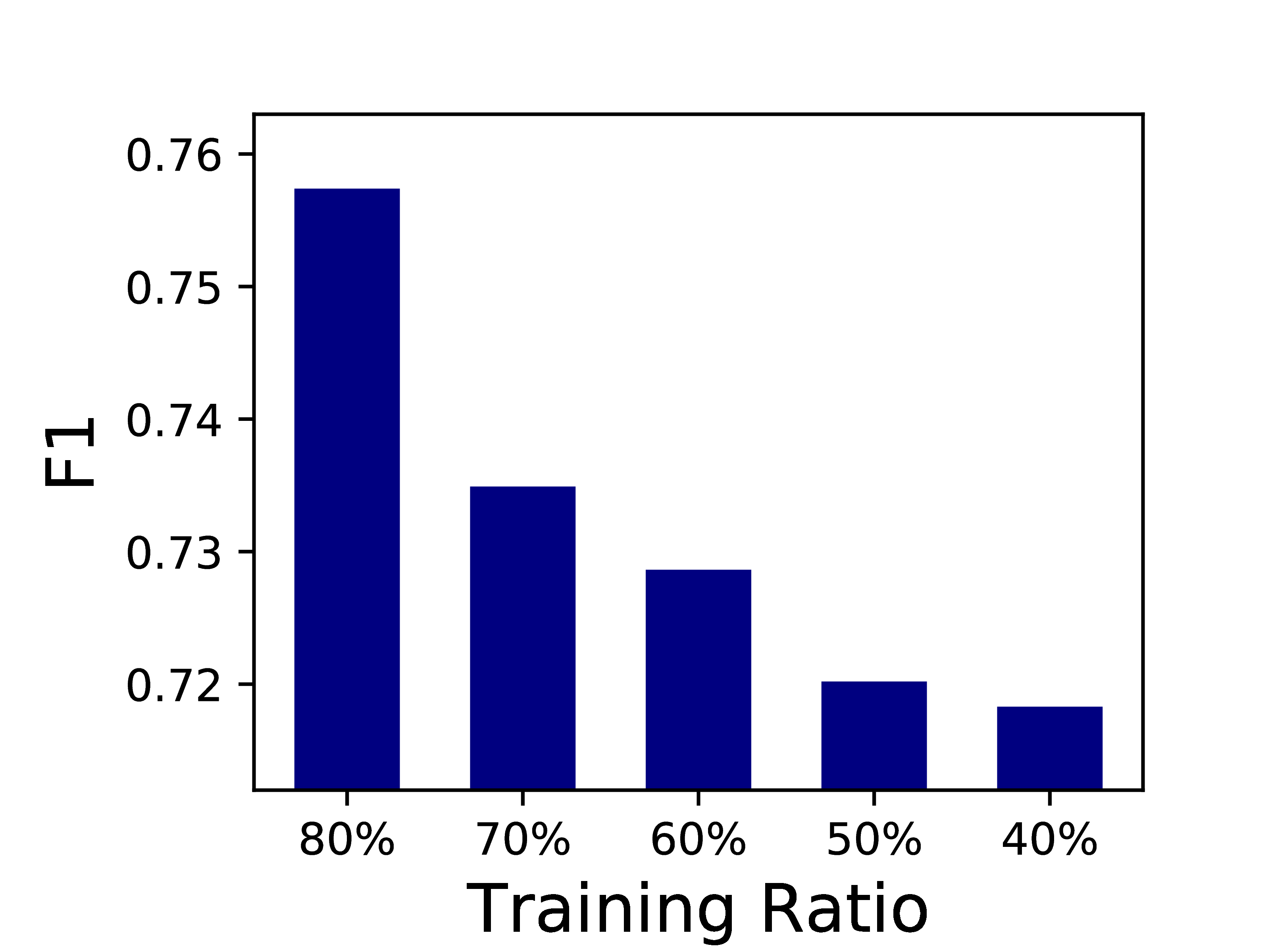}
\label{F1_testratio}}
\hspace{-0.27in}
\subfigure[The AUC performance]{
\includegraphics[width=1.7in]{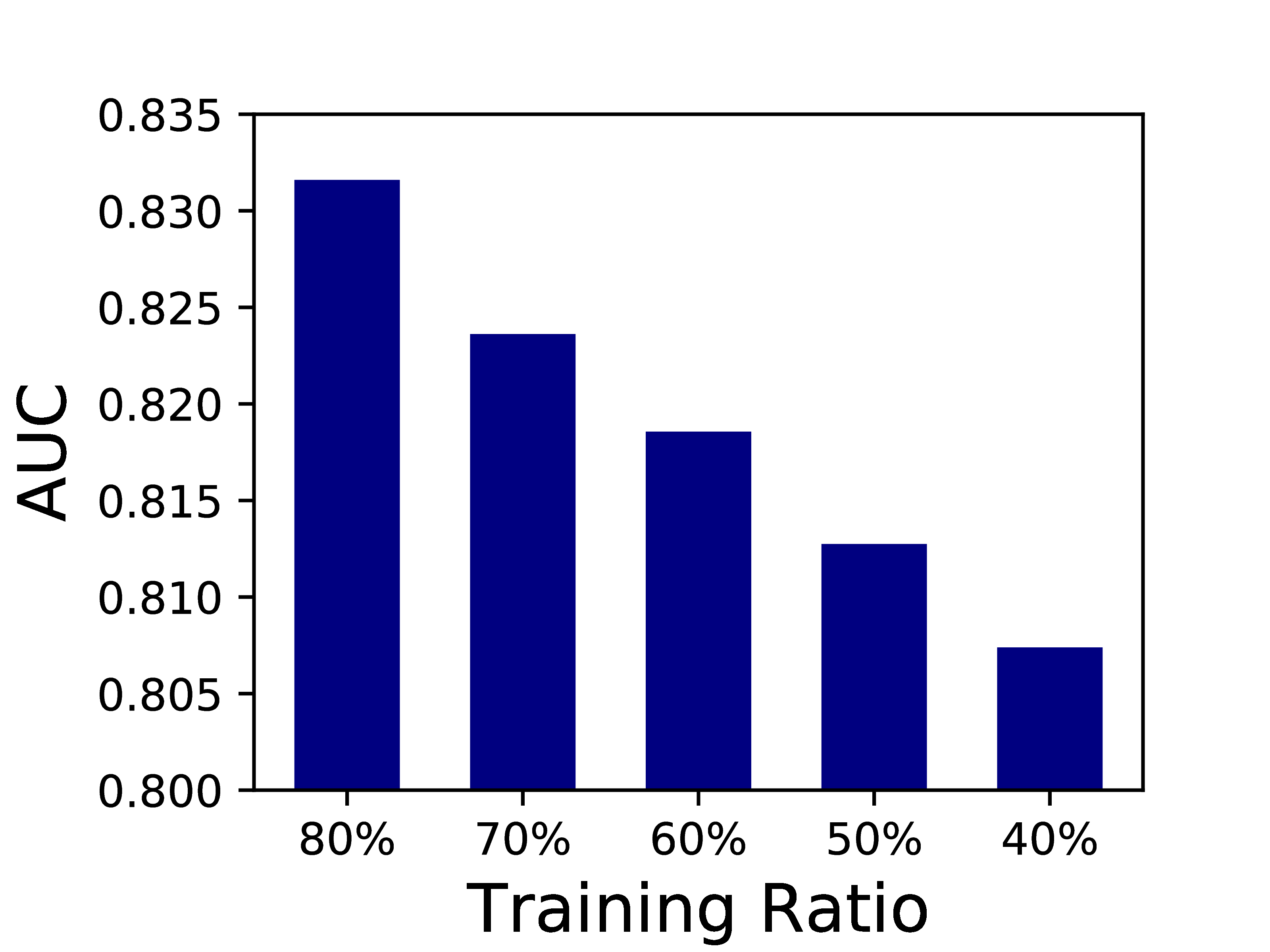}
\label{AUC_testratio}}
\caption{The performance of APJFNN at different train/test split.}
\end{figure}

\noindent\textbf{$\bullet$ The Robustness on Different Data Split.} To observe how our model performs at different train/test split, we randomly selected 80\%, 70\%, 60\%, 50\%, 40\% of the dataset as training set, another 10\% for tuning the parameters, and the rest part as testing set~\footnote{The numbers of samples in training/validation/testing set were 20,446/2,556/2,556; 17,891/2,556/5,111; 15,335/2,556/7,667; 12,779/2,556/10,223 and 10,223/2,556/12,779 respectively.}. The results are shown in Figure \ref{F1_testratio}, \ref{AUC_testratio}. We can observe that the overall performance of our model is relatively stable, while it gets better as the training data increases. Indeed, the improvements of the best performance compared with the worst one are only 5.44\% and 2.99\% for two metrics respectively. Furthermore, we find that our model with 60\% of data for training has already outperforms all the baselines methods, which use 80\% of the data for training. The results clearly validate the robustness of our model in terms of training scalability.

\begin{figure}\centering
	\vspace{-2mm}
	\includegraphics[width=3.0in]{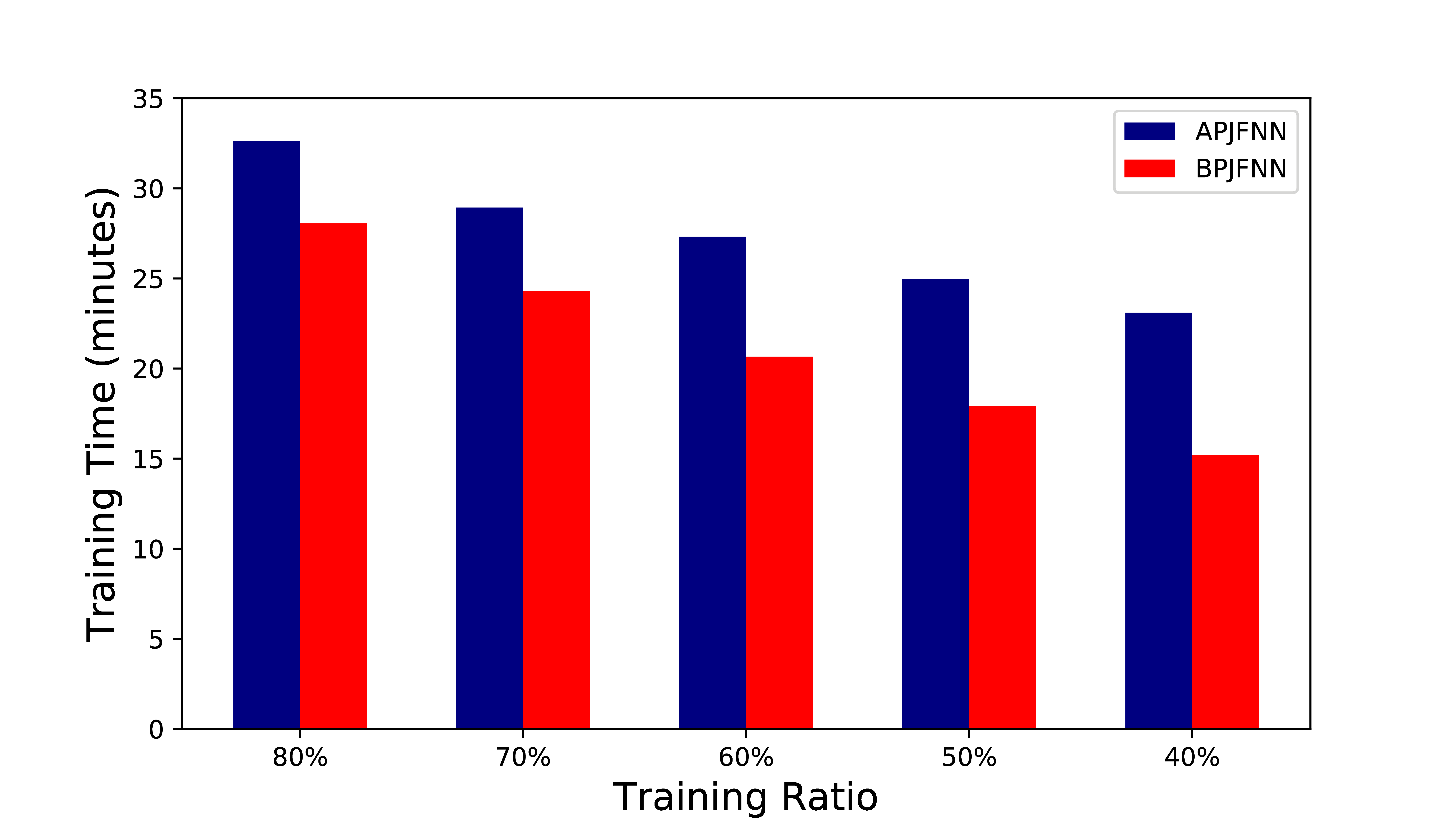}
	\caption{The training efficiency of APJFNN at different train/test split.}
	\vspace{-2mm}
	\label{efficiency}
\end{figure}

\begin{figure*}[t]
  \includegraphics[width=6.9in]{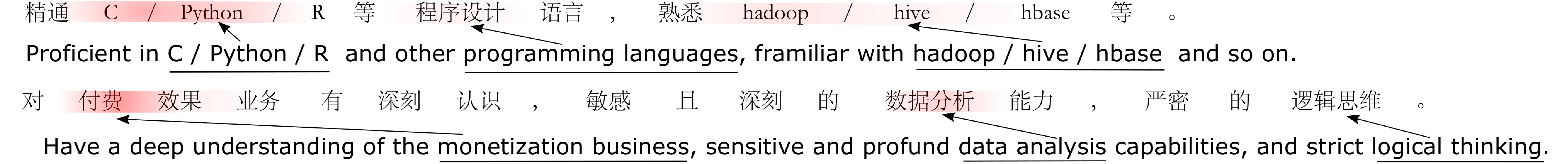}
  \caption{Two examples for demonstrating the advantage of Attention $\alpha$ in capturing the informed part of the ability requirement sentence.}
  \label{att0}
  \vspace{-3mm}
\end{figure*}

\begin{figure}[t]
  \includegraphics[width=3.4in]{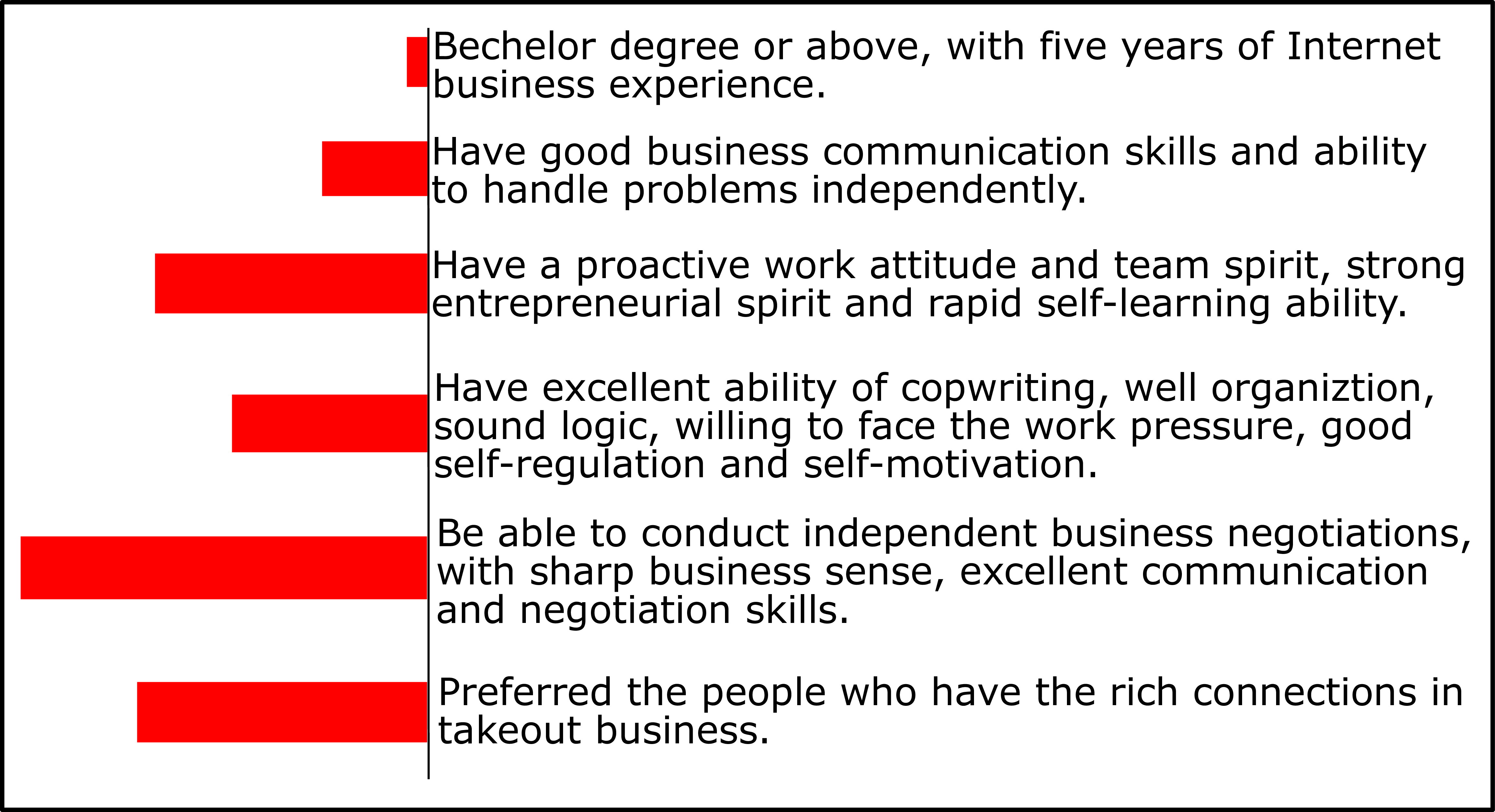}
  \caption{An example for demonstrating the advantage of Attention $\beta$ in measuring the importance of the each ability requirement among all the job needs. The left bar charts denote the distribution of $\beta$ over all requirements. }
  \vspace{-4.5mm}
  \label{att1}
\end{figure}

\noindent\textbf{$\bullet$ Computational Efficiency.} Here we evaluate the computational efficiency of our model APJFNN. Specifically, all of our experiments were conducted on a server with 2-core CPU@2.40GHz, 160GB RAM, and a Tesla K40m GPU. First, we present the training time of different data split. As shown in Figure~\ref{efficiency}, we observe the training time of our model does not increase dramatically with the increase of training data. Although our model is relatively slower than the BPJFNN, however, it can achieve the better performance as presented in the Table~\ref{real-world}. Moreover, after the training process, the average cost of each instances in testing set is 13.46ms. It clearly validate that our model can be effectively used in the real world recruitment analysis system.

\subsection{Case Study}
With the proposed attention strategies, we target at not only improving the matching performance, but also enhancing the interpretability of matching results. To that end, in this subsection, we will illustrate the matching results in three different levels by visualizing the attention results.

\vspace{1mm}
\noindent $\diamond$ \textbf{~Word-level: Capturing the key phrases from the sentences of job requirement.}
\vspace{1mm}

Firstly, we would like to evaluate whether our APJFNN model could reveal the word-level key phrase from long sentences in job requirements. The corresponding case study is shown in Figure~\ref{att0}, in which some words (in Chinese) are highlighted as \emph{key phrases}, and their darkness correlated to the value of attention $\alpha$.

According to the results, it is unsurprising that the crucial skills are highlighted compared with common words. Furthermore, in the same requirement, different abilities may have different importance. For instance, In the requirement in line 1, which is technique-related, \emph{C/Python/R} could be more important than \emph{Hadoop}, which might be due to the different degrees (``proficient'' v.s. ``familiar''). Similarly, for the product-related requirement in line 2, more detailed skills are more important, e.g., data analysis compared with logical thinking.

\vspace{1mm}
\noindent $\diamond$ \textbf{~Ability-level: Measuring the different importance among all abilities.}
\vspace{1mm}

Secondly, we would like to evaluate whether APJFNN could highlight the most critical abilities. The corresponding case study is shown in Figure~\ref{att1}, in which histogram indicates the importance of each ability, i.e., the distribution of attention $\beta$.

From the figure, striking contrast can be observed among the 6 abilities, in which the \emph{bachelor degree} with the lowest significance is usually treated as the basic requirement. Correspondingly, the ability of \emph{independent business negotiation} could be quite beneficial in practice, which leads to the highest significance. In other words, the importance of abilities could be measured by the \emph{scarcity}, as most candidates have the bachelor degree, but only a few of them could execute business negotiation independently.

\begin{figure*}
  \includegraphics[width=6.8in]{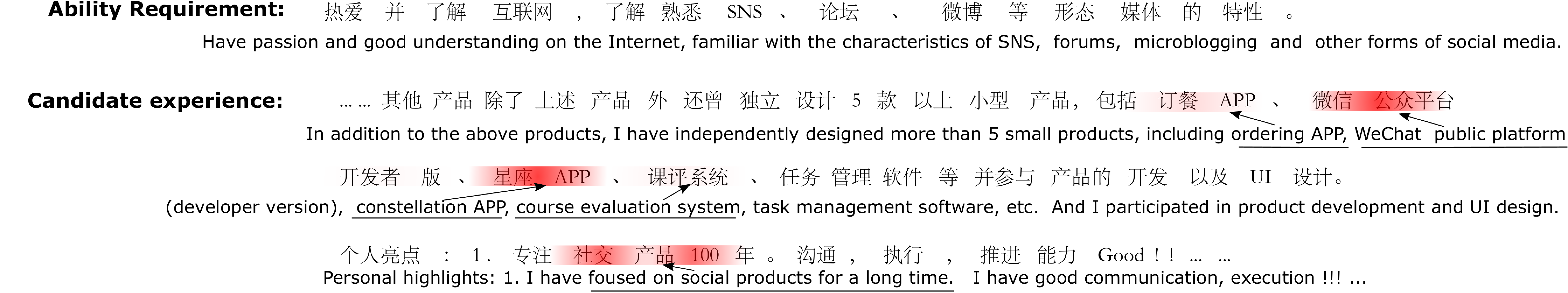}
  \caption{An example for demonstrating the advantage of Attention $\gamma$ in capturing the ability-aware informed part from the experience of candidate.}
  \label{att2}
\end{figure*}

\vspace{1mm}
\noindent $\diamond$ \textbf{~Matching-level: Understanding the matching between job requirements and candidate experiences.}
\vspace{1mm}

At last, we would like to evaluate how APJFNN model could guide the matching between requirements and experiences. The corresponding case study is shown in Figure~\ref{att2}, in which darkness is also correlated to the importance of experience with considering the different job requirements, i.e., the attention value of $\gamma$.

Definitely, we find that those key phrases which could satisfy the requirements are highlighted, e.g., \emph{WeChat public platform} and \emph{focus on social products} for the requirement \emph{SNS, forums}. Also, we realize that the ``importance'' here indeed indicates the degree of satisfying the requirements. For instance, the phrase \emph{WeChat public platform} (a famous SNS in China) is darker than \emph{ordering APP}, since the former one is strongly related to the \emph{SNS} requirement, but the latter one is only a rough matching. Thus, this case study also proves that our APJFNN method could provide good interpretability for Person-Job Fit task, since key clues for matching the job requirements and candidate experience can be highlighted.

\vspace{-2mm}
\section{Conclusions}
In this paper, we proposed a novel end-to-end Ability-aware Person-Job Fit Neural Network (APJFNN) model, which has a goal of reducing the dependence on manual labour and can provide better interpretation about the fitting results. The key idea is to exploit the rich information available at abundant historical job application data. Specifically, we first proposed a word-level semantic representation for both job requirements and job seekers' experiences based on Recurrent Neural Network (RNN). Then, four hierarchical ability-aware attention strategies were designed to measure the different importance of job requirements for semantic representation, as well as measuring the different contribution of each job experience to a specific ability requirement. Finally, extensive experiments conducted on a large-scale real-world data set clearly validate the effectiveness and interpretability of our APJFNN framework compared with several baselines.

\vspace{-2.5mm}


\section{ACKNOWLEDGMENTS}
This work was partially supported by grants from the National Natural Science Foundation of China (Grant No.91746301, U1605251, 61703386).

\vspace{-1mm}

\appendix 
\section*{APPENDIX}

Admittedly, while the accuracy of the algorithm is essential, another paramount issue, which needs to be paid attention to, is ensuring the fairness of the algorithm and empowering the correct values of intelligent recruitment system. In recent years, it has been received extensive attention from academics and the media~\cite{amazonbias}. As for machine learning based algorithms, avoiding bias in training data is necessary for their fairness, such as the significant difference in employment ratio of women to men. Unfortunately, for many existing recruitment practices in our real life, the prejudices seem hard to be completely avoid. For example, according to a recent report~\cite{tokyomedical}, doctor has long been a male bastion of the Tokyo Medical University, where they confessed to marking down the test scores of female applications to keep the ratio of women in each class below 30\%. 

So, in the construction of the intelligent recruitment system, one of the questions that must be answered is that if we already have a dataset with potential value discrepancy, how can we avoid further misleading the algorithm? Intuitively, if the data with gender bias are used for training machine learning models of intelligent recruitment, \emph{Gender} would be regarded as a dominant feature based on the commonly feature engineering, since whether the Chi-squared test result, information gain or correlation coefficient score indicate that it has a significant correlation with the recruitment result. Therefore, \emph{Gender} feature is seen as a potential factor affecting the values of the machine learning algorithm. In our conjecture, we should not add \emph{Gender} feature to train the model.
=
In order to confirm our conjecture, here we adjust equation~\ref{D_eq} to: 
\begin{equation*}
\begin{split}
D &= tanh(W_d[o;g^J;g^R;g^J-g^R]+b_d), \\
\widetilde{y} &= {Sigmoid}(W_yD+b_y),
\end{split}
\end{equation*}
where $o$ is the \emph{Gender} feature. And we evaluate on a semi-synthetic data based on a real-world recruitment system. First of all, we constructed a ``balanced dataset'' in terms of gender.  Specifically, we randomly selected 5,678 successful job applications (positive instances) from the recruitment records of historical job postings, where half of them are female candidates. Then, for each of the job postings, we also randomly selected the same number of failed job applications (negative instances). In particular, both successful and failed applications satisfy that the numbers of male and female candidates are equal.
Next, in the model validation step, we randomly selected 80\% of the dataset as training data, another 10\% for tuning the parameters, and the last 10\% as test data to validate the performance and robustness. As same time, in order to simulate the possible unfairness scenario in the recruitment system, we randomly labeled 50\% female successful applications as negative, and labeled 50\% male failed applications as positive ones, in the \textbf{training set} and \textbf{validation set}.  After the manual construction, in both training and validation sets, the success rates of male and female candidates become 75\% and 25\%, respectively. Note that, we did not change the labels in \textbf{test set}, where has the same cutoff ratio as "balance dataset" for both women and men to ensure it has the correct values.

Table~\ref{tab:semi-synthetic} shows the performance on the validation set and testing set of the semi-synthetic data. Clearly, we observe that with Gender feature, each model in validation set has better performance since validation set has similar distribution with training set. However, in other words, those models have unfortunately learned the value bias that existed therein. In contrast, we realize that all the models perform better without using gender information on the testing set, which demonstrates that the models can avoid value deviation from the training data to a great extent without leveraging the Gender information. Therefore, we can conclude that when historical recruitment dataset contains the bias of data distribution, such as gender discrimination, we should not use the corresponding features to train the model, thus avoiding algorithm to produce value deviations like humans.

\begin{table}[t]
\caption{The performance of APJFNN and baselines on semi-synthetic data.}
\resizebox{\columnwidth}{!}{%
\begin{tabular}{ l | c | c | c | c | c | c | c | c | c | c | c  }
\hline \hline
  
  \multicolumn{2}{l|}{Features} & \multicolumn{5}{c}{Without gender feature} & \multicolumn{5}{|c}{With gender feature} \\  \hline 
  
  Methods & Datasets & Accuracy & Precision & Recall & F1 & AUC & Accuracy & Precision & Recall & F1 & AUC  \\ \hline \hline 
  
  \multirow{2}{*}{LR} & Validation set & 0.5122 & 0.5126 & 0.4957 & 0.5040 & 0.5348 & 0.6783 & 0.6758 & 0.6852 & 0.6805 & 0.7063 \\ 
   & Testing set & 0.5855 & 0.5913 & 0.5913 & 0.5913 & 0.6093 & 0.5203 & 0.5281 & 0.5061 & 0.5169 & 0.5693 \\  \hline 
  
  \multirow{2}{*}{AB} & Validation set & 0.5713 & 0.5724 & 0.5635 & 0.5679 & 0.5847 &  0.7217 & 0.7040 & 0.7652 & 0.7333 & 0.7882 \\
   & Testing set & 0.6402 & 0.6567 & 0.6087 & 0.6318 & 0.6770 & 0.5459 & 0.5549 & 0.5270 & 0.5406 & 0.6274 \\ \hline 
   
   \multirow{2}{*}{DT} & Validation set & 0.5870 & 0.6179 & 0.4557 & 0.5245 & 0.5951 &  0.7261 & 0.7167 & 0.7478 & 0.7319 & 0.7744 \\
   & Testing set & 0.6711 & 0.7349 & 0.5496 & 0.6289 & 0.6807 & 0.5079 & 0.5159 & 0.4800 & 0.4973 & 0.5701 \\  \hline 
   
   \multirow{2}{*}{RF} & Validation set & 0.5991 & 0.6096 & 0.5513 & 0.5790 & 0.6118 &  0.7148 & 0.6939 & 0.7687 & 0.7294 & 0.7531 \\
   & Testing set & 0.6279 & 0.6527 & 0.5687 & 0.6078 & 0.6857 & 0.5141 & 0.5211 & 0.5165 & 0.5188 & 0.5807 \\ \hline 
   
   \multirow{2}{*}{GBDT} & Validation set & 0.5913 & 0.5953 & 0.5704 & 0.5826 & 0.6290 &  0.7200 & 0.7030 & 0.7617 & 0.7312 & 0.7945 \\
   & Testing set & 0.6896 & 0.7069 & 0.6626 & 0.6840 & 0.7436 & 0.5194 & 0.5271 & 0.5078 & 0.5173 & 0.6208 \\ \hline
   
   \multirow{2}{*}{LR(with word2vec)} & Validation set & 0.5652 & 0.5693 & 0.5357 & 0.5520 & 0.5985 & 0.7113 & 0.6989 & 0.7426 & 0.7201 & 0.7625 \\
   & Testing set & 0.5873 & 0.6011 & 0.5530 & 0.5761 & 0.6140 & 0.5079 & 0.5150 & 0.5078 & 0.5114 & 0.5642 \\ \hline 
   
   \multirow{2}{*}{AB(with word2vec)} & Validation set & 0.5626 & 0.5655 & 0.5409 & 0.5529 & 0.5780 & 0.7217 & 0.7121 & 0.7443 & 0.7280 & 0.7685 \\
   & Testing set & 0.5540 & 0.5647 & 0.5235 & 0.5433 & 0.5860 & 0.5256 & 0.5322 & 0.5322 & 0.5322 & 0.5565 \\ \hline 
   
   \multirow{2}{*}{DT(with word2vec)} & Validation set & 0.5313 & 0.5304 & 0.5461 & 0.5381 & 0.5577 & 0.7243 & 0.7067 & 0.7670 & 0.7356 & 0.7435 \\
   & Testing set & 0.5502 & 0.5534 & 0.5861 & 0.5693 & 0.5853 & 0.4929 & 0.5000 & 0.5009 & 0.5004 & 0.5340 \\ \hline 
   
   \multirow{2}{*}{RF(with word2vec)} & Validation set & 0.5565 & 0.5610 & 0.5200 & 0.5397 & 0.5756 & 0.6991 & 0.6856 & 0.7357 & 0.7097 & 0.7332 \\
   & Testing set & 0.5847 & 0.6057 & 0.5183 & 0.5586 & 0.6301 & 0.5212 & 0.5282 & 0.5217 & 0.5249 & 0.5387 \\ \hline 
   
   \multirow{2}{*}{GBDT(with word2vec)} & Validation set & 0.5809 & 0.5841 & 0.5617 & 0.5727 & 0.5983 & 0.7157 & 0.7033 & 0.7461 & 0.7241 & 0.7687 \\
   & Testing set & 0.5970 & 0.6105 & 0.5670 & 0.5979 & 0.6317 & 0.5088 & 0.5157 & 0.5130 & 0.5144 & 0.5587 \\ \hline 
   
   \multirow{2}{*}{PJFNN-RNN} & Validation set & 0.5974 & 0.6261 & 0.4835 & 0.5456 & 0.6443 & 0.7183 & 0.6865 & 0.8035 & 0.7404 & 0.7858 \\
   & Testing set & 0.6296 & 0.6824 & 0.5043 & 0.5800 & 0.7034 & 0.5397 & 0.5425 & 0.5878 & 0.5643 & 0.6178 \\ \hline 
   
   \multirow{2}{*}{APJFNN} & Validation set & 0.6191 & 0.6179 & 0.6243 & 0.6211 & 0.6681 & 0.7157 & 0.6649 & 0.8696 & 0.7536 & 0.8091 \\
   & Testing set & 0.7425 & 0.7386 & 0.7617 & 0.7500 & 0.8036 & 0.5917 & 0.5780 & 0.7217 & 0.6419 & 0.6444 \\

  \hline\hline
\end{tabular}}
\label{tab:semi-synthetic}
\end{table}

\bibliographystyle{ACM-Reference-Format}
\bibliography{reference}

\end{document}